\documentclass{article}
\usepackage[dvipsnames]{xcolor}
\usepackage{pifont}
\usepackage{colortbl}
\usepackage{comment}
\usepackage{graphicx}
\usepackage{multirow}
\usepackage{booktabs}
\usepackage{amssymb}
\usepackage{amsmath}
\usepackage{wrapfig} %

\usepackage{enumitem}\setlist{nosep,noitemsep}
\usepackage{subcaption}

\newcommand{\cross}{\textcolor{Salmon}{\textbf{\texttimes}}}
\newcommand{\cmark}{\textcolor{YellowGreen}{\textbf{\checkmark}}}

\usepackage[preprint]{corl_2025} %
\usepackage{cleveref}
\usepackage{authblk}

\title{Splatting Physical Scenes: End-to-End Real-to-Sim from Imperfect Robot Data}

\author[*,1,2]{Ben Moran}
\author[*,1,3]{Mauro Comi}
\author[1]{Arunkumar Byravan}
\author[1]{Steven Bohez}
\author[1]{Tom Erez}
\author[2]{\authorcr Zhibin Li}
\author[1]{Leonard Hasenclever}
\affil[1]{Google DeepMind}
\affil[2]{University College London}
\affil[3]{University of Bristol}

\begin{document}

\maketitle

\renewcommand{\thefootnote}{\fnsymbol{footnote}}
\footnotetext[1]{Equal contribution.}
\renewcommand{\thefootnote}{\arabic{footnote}}
\let\thefootnote\relax\footnotetext{Correspondence to \texttt{benmoran@google.com}}
\begin{figure}[h!]
    \centerline{\includegraphics[width=0.97\columnwidth]{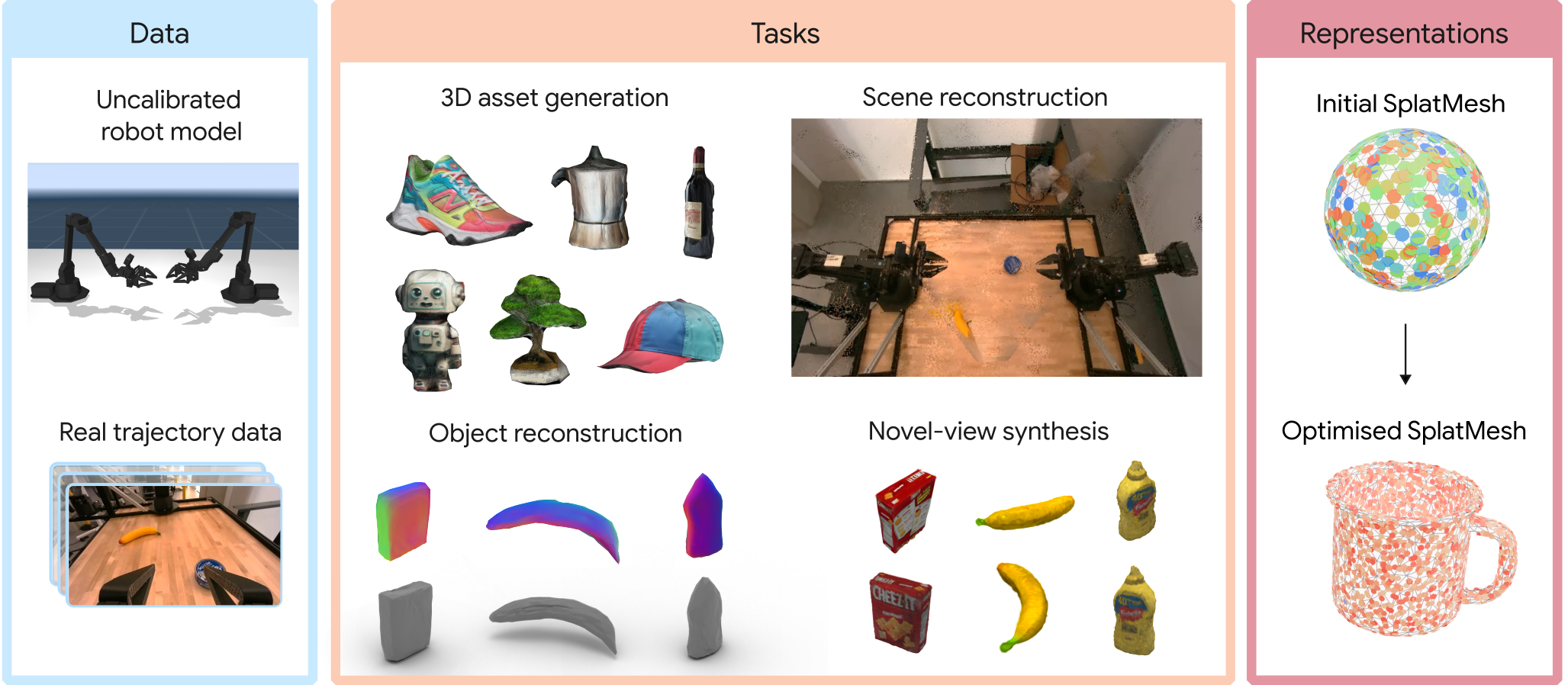}}
    \caption{Overview of our framework. Starting from an uncalibrated robot and real-world sensor data (RGB images and proprioception), we perform robot calibration, reconstruct both the geometry and appearance of scene elements.}
    \label{fig:fig1}
\end{figure}

\begin{abstract}
Creating accurate, physical simulations directly from real-world robot motion holds great value for safe, scalable, and affordable robot learning, yet remains exceptionally challenging. Real robot data suffers from occlusions, noisy camera poses, dynamic scene elements, which hinder the creation of geometrically accurate and photorealistic digital twins of unseen objects.
We introduce a novel real-to-sim framework tackling all these challenges at once. Our key insight is a hybrid scene representation merging the photorealistic rendering of 3D Gaussian Splatting with explicit object meshes suitable for physics simulation \textit{within a single representation}. We propose an end-to-end optimization pipeline that leverages differentiable rendering and differentiable physics within MuJoCo to jointly refine all scene components – from object geometry and appearance to robot poses and physical parameters – directly from raw and imprecise robot trajectories. This unified optimization allows us to simultaneously achieve high-fidelity object mesh reconstruction, generate photorealistic novel views, and perform annotation-free robot pose calibration.
We demonstrate the effectiveness of our approach both in simulation and on challenging real-world sequences using an ALOHA 2 bi-manual manipulator, enabling more practical and robust real-to-simulation pipelines.
\end{abstract}


\section{Introduction}

Creating physically accurate and visually realistic simulations directly from real-world robot interactions is crucial for scalable robotics, yet bridging the visual and physical sim-to-real gap remains a major hurdle, especially with imperfect data. While recent advances like Neural Radiance Fields (NeRF) \cite{mildenhall2020nerf} and 3D Gaussian Splatting (3DGS) \cite{3dgs} generate high-quality photorealistic novel views \cite{mildenhall2020nerf, 3dgs}, they face significant challenges in dynamic robotic settings: they are sensitive to noisy camera poses common in real trajectories, and produce representations ill-suited for direct use in physics simulators like MuJoCo \cite{todorov2012mujoco}. Extracting usable simulation assets often requires laborious post-processing \cite{Dogaru2024-dz} or separate geometry estimation pipelines, breaking the link between visual input and physical behaviour. This disconnect critically limits the automated creation of high-fidelity digital twins for robot learning and planning. 

We argue that overcoming this requires tackling scene appearance reconstruction, object geometry extraction, and robot/camera calibration jointly within a single, end-to-end differentiable framework. This paper introduces such a framework, enabling the learning of explicit simulation scenes directly from imperfect, dynamic robot trajectory data. Our key innovation is the development of a single representation that combines \textit{SplatMesh}, a hybrid scene representation that tightly couples 3DGS for appearance with explicit mesh geometry, with differentiable physics states. 

Our end-to-end optimization pipeline leverages differentiable rendering and differentiable physics using MuJoCo MJX \cite{brax2021github}. This allows visual discrepancies to directly propagate gradients back through the entire system, simultaneously refining not only the 3DGS appearance but also the underlying mesh geometry, estimated robot poses, and camera parameters. This unified approach eliminates the need for separate processing steps and leverages visual feedback to ground the physical reconstruction, proving effective even with imperfect real robot platforms. We demonstrate the effectiveness of our framework on object reconstruction with the low-cost ALOHA bi-manual manipulator using \textbf{only onboard RGB sensors} and a pre-existing, imperfect robot model. 

Our contributions are as follows:
\begin{itemize}
    \item \emph{End-to-End Real-to-Sim Pipeline:} A fully differentiable framework jointly optimizing scene appearance, object geometry, robot poses, and camera parameters directly from raw RGB sequences.
    \item \emph{Scene Reconstruction from Imperfect Data:} Demonstrated reconstruction of dynamic robot scenes, including novel objects, from noisy monocular trajectories captured by real robots.
    \item \emph{Simulation-Ready Asset Generation:} Controllable reconstruction of object meshes suitable for direct integration into physics engines, driven by visual consistency.
\end{itemize}

\section{Related work}

\subsection{3D scene reconstruction}
Radiance fields represent a 3D scene in terms of a rendering function which determines the view-dependent color and opacity of each point $(x,y,z)$ in the 3D space of the scene.  Evaluating this along the 3D rays for each pixel enables the scene to be consistently rendered from arbitrary camera viewpoints. This function can be parameterized in different ways - implicitly via Neural Networks as in such as Neural Radiance Fields \cite{mildenhall2020nerf}, or explicitly as in 3D Gaussian Splatting (3DGS) \cite{3dgs}. %

The key assumption of these methods is knowledge of the 3D structure of the generation process of the multi-view image data, usually by requiring scene to be static across the image views. The camera parameters must also be known precisely, typically by preprocessing with COLMAP \cite{schoenberger2016sfm,schoenberger2016mvs} which also performs best on static scenes.

\begin{table}[tbp]
\centering
\caption{Feature Comparison of 3DGS robotics simulators }
\renewcommand*{\arraystretch}{1.}
\label{tab:feature_comparison}

\scalebox{0.77}{
    \begin{tabular}{l|c|c|c|c|c|c|c}
    \toprule
    \textbf{Feature} & \textbf{DRRobot} & \textbf{RL-GSBridge} & \textbf{RoboGSim} & \textbf{RoboStudio}  & \textbf{1SR2S} & \textbf{SplatSim}& \textbf{Ours} \\
     & \cite{Liu2024-mk-drrobot} & \cite{wu2024rlgsbridge} & \cite{Li2024-bm-RoboGSim} &  \cite{Lou2024-lo-robostudio} & \cite{Zhu2024-zk-tencent} & \cite{qureshi2024splatsim} & \\
    
    \midrule
    
    Render robot kinematics & \cmark & \cmark & \cmark & \cmark & \cmark & \cmark & \cmark \\
    
    Learn novel object mesh & \cross & \cross & \cross & \cross & \cmark & \cross & \cmark \\
    
    Physics simulation & \cross & \cross & \cross & \cross & \cmark & \cmark & \cmark \\
    
    Differentiable Physics & \cross & \cross & \cross & \cross & \cmark & \cross & \cmark \\
    
    Learning from dynamic scenes & \cross & \cross & \cross & \cross & \cross & \cross & \cmark \\
    
    3D asset generation & \cross & \cross & \cross & \cross & \cross & \cross & \cmark \\
    
    \bottomrule
    \end{tabular}
    }
\end{table}

\subsection{Real2Sim robotics with radiance fields}

Beyond the computer graphics community, there has been considerable research into radiance fields applied to robotics \cite{Byravan2022-tp, rapidsim-10543811, rosinol2022nerfslam}. Several recent works obtain 3DGS representations able to render a robot arm and its articulated poses, such as \cite{Liu2024-mk-drrobot, wu2024rlgsbridge, Li2024-bm-RoboGSim, Lou2024-lo-robostudio, Zhu2024-zk-tencent}. A 3DGS representation of the robot is learned either from its CAD model or from manually collected posed photographs, and the Gaussians then segmented according to the robot model.  A learned mapping between the kinematic configurations and part poses is used to render the robot in arbitrary joint configurations.  Articulated 3DGS models must also be learned for each simulated interaction object.  Such models can be applied to trajectory tracking via inverse video \cite{Liu2024-mk-drrobot}, sim-to-real policy learning in visually realistic simulations (e.g. \cite{Li2024-bm-RoboGSim}), or physical property estimation \cite{Zhu2024-zk-tencent}.  We compare the features of these works with ours in table \ref{tab:feature_comparison}. We additionally focus on the ability to fit models to imprecise low-cost robots without separate data collection steps.
\subsection{Object-based scene decomposition}

Many applications require compositionally structured scenes with objects. Pre-trained instance segmentation models enable masking the objects of interest, before fitting the masked images separately with local radiance fields.  For example, \cite{kong2023vmap} uses the Detic model \cite{zhou2022detic} combined with NeRF, while \cite{Ye2023-qp} combines SAM \cite{kirillov2023segment} with 3DGS, and \cite{Dogaru2024-dz} considers various methods together with CropFormer \cite{qi2023highqualityentitysegmentation}.

\section{Problem setting}

Given a reasonable but inaccurate simulation of our robot, and samples of real observation data, we wish to recover an accurate simulation of our scene, including new scene geometry.

\subsection{Fitting models with prediction errors}
For any simulation of a system we can always consider ``prediction error'': how well do the simulator predicted observations agree with data from the real world? 

Formally, given a multi-modal robot observation $Y$ composed of $I$ modality specific components, $Y=(y_1, \dots, y_I)$, we choose divergence functions $d_i(\cdots, \cdots)$ and weights $\beta_i \in \mathbb{R}$.  Our idealized simulator is a model that generates observations from physics states $s \in \mathcal{S}$: $Y' = m(s)$. As the physics state may be large and is not necessarily differentiable, we will further consider perturbations from a base state $s' = f(s_0, \theta)$ with respect to a selected parameter set $\theta \in \Theta$.  We can then finally define our loss as the weighted sum of the divergence terms over each observation component:  $$L(\theta) = \sum_{i\in I} \beta_i d_i(Y, m(f(s_0, \theta))).$$

\subsection{Real-world constraints in robotics}

This high-level scheme is very general, so to ground our investigation, we consider a realistic real-to-simulation problem: access to a simulator with reasonably accurate geometry and kinematics, but not perfect system identification or camera calibration. The scene also contains an unmodelled novel object, and must be learned from on-board robot sensors without additional data collection. We consider the ALOHA2 low-cost bi-manual tabletop manipulation platform \cite{zhao2023ALOHA}\cite{aloha2team2024aloha2enhancedlowcost}, together with its associated open-source MuJoCo model \cite{menagerie2022github}. This limits us to four RGB cameras, two fixed and two mounted on the moving wrists. The robot has 6 degrees of freedom in each arm and 1 in each gripper, with Dynamixel actuators.

This setting poses obstacles for standard 3DGS data collection and scene modelling: a small number of cameras enables only a constrained range of viewpoints; the motion of the robot arms makes the scenes dynamic; camera position estimates are noisy due to timing, backlash, imperfect encoder calibration etc. These challenging conditions mean several popular pipeline components are not applicable here. We tested COLMAP on both masked object images, and on the full image trajectories, but could not obtain coherent estimates across the cameras. Moreover, as shown in the Appendix, object segmentation models like SAM2 can provide good segmentations for semantic scene objects, but are not effective for segmenting elements like the robot body that have little texture, lack clear semantic descriptions, and are easily confounded with similar distractor elements in the scene background. To evaluate the recovered simulation, we perform (1) calibration of camera extrinsics and robot pose, (2) novel view synthesis, and (3) novel object geometry reconstruction.

\section{Method}
\subsection{System overview}
\begin{figure}[t]
    \centerline{\includegraphics[width=0.95\columnwidth]{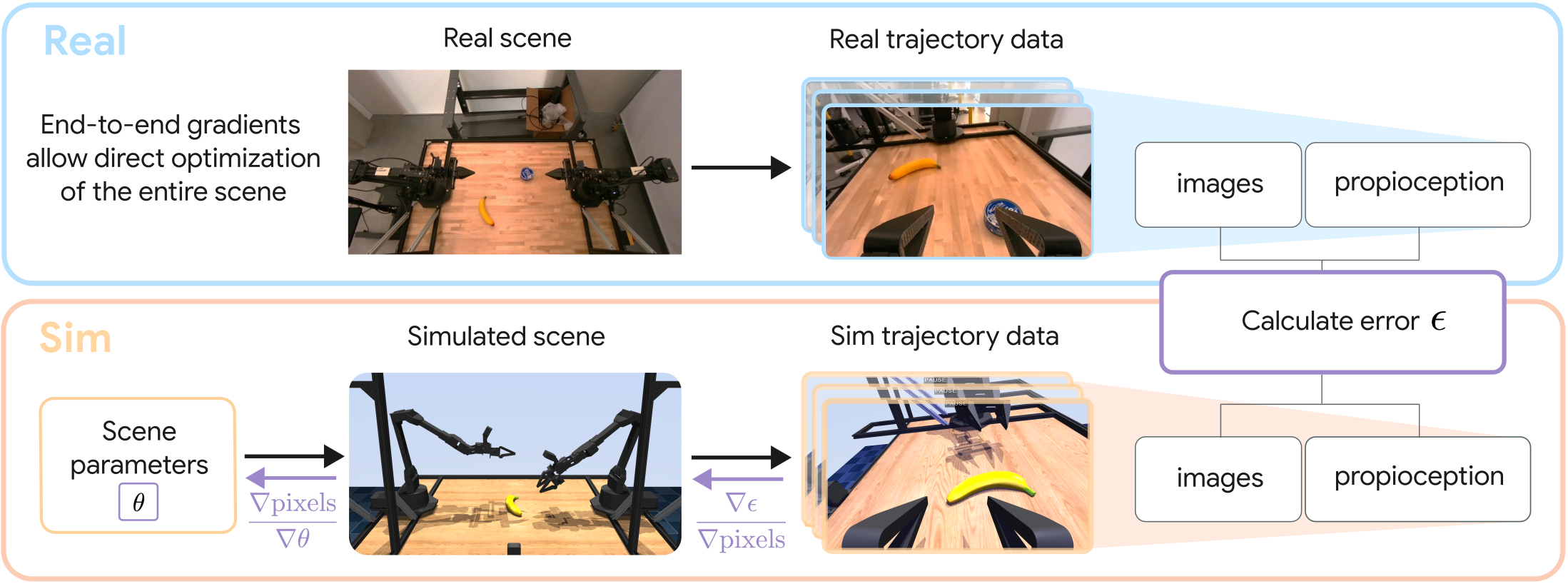}}
    \caption{Fitting with differentiable rendering. We optimize the scene parameters $\theta$, consisting of object vertices, 3D Gaussian parameters, camera poses and robot joint angles, uniquely using real-world data acquired by robot sensors.}
    \label{fig: example}
\end{figure}

We propose a general framework that solves a diverse range of tasks in a real, low-cost, bi-manual robot setting via end-to-end optimization of all the components in our scene. To achieve our goal, we implement a prediction error optimization scheme using automatic differentiation and GPU acceleration. Specifically, we first collect real RGB images and robot states from recorded robot trajectories to build a model of our scene. For this purpose, we propose a novel scene representation that enables end-to-end optimization of all the represented elements. Then, we optimize the required components via this robot trajectory data alone, using differentiable physics simulation and differentiable rendering.  Please refer to the Appendix for full details.

\subsection{Scene representation}
\subsubsection{SplatMesh}
\label{sec:splatmesh}

We represent scene objects using \textit{SplatMesh}, a hybrid representation combining a triangle mesh for geometry with 3D Gaussians for appearance (Fig. \ref{fig:fig1}). 3D Gaussians are constrained to lie on the surface of the mesh faces, and their orientations transform as the underlying mesh moves. Decoupling visual and geometry information allows flexibility to learn the appearance, pose, and/or shapes of each element, or to treat these as fixed by the original model. 

To optimize the geometry, we deform the vertices while preserving the underlying connectivity. This approach maintains a consistent mesh topology, ensuring a fixed and controllable number of vertices and faces. The optimization of the explicit underlying geometry offers two key advantages. First, it enables the direct incorporation of mesh regularization terms into the optimization objective, promoting desirable properties such as smoothness and mesh uniformity. Second, it provides precise control over mesh complexity, resulting in computationally efficient simulations of the reconstructions. 

The mean $\mathbf{\mu}$ of each Gaussian is initialized using a weighted barycentric coordinate approach. Given a face defined by the vertices $\mathbf{v_1}, \mathbf{v_2}, \mathbf{v_3} \in \mathbb{R}^3$, we randomly sample barycentric weights (implicitly summing to 1) to determine the Gaussians positions:
$$ \boldsymbol{\mu} = w_1 \cdot \mathbf{v}_1 + w_2 \cdot \mathbf{v}_2 + w_3 \cdot \mathbf{v}_3.$$
Each 3D Gaussian is further parameterized by a covariance $\Sigma \in \mathbb{R}^{3\times3}$, view-dependent spherical harmonics coefficients representing color, and an opacity term $o \in [0, 1]$. Object appearance is rendered through the differentiable rasterization of the 3D Gaussians. All SplatMesh parameters, encompassing both appearance and geometry, can be optimized via supervision from RGB images thanks to the differentiable pipeline connecting geometry, Gaussians, and the final rendering.

\subsection{End-to-end optimization}
Given a set of input images and an initial scene composed of a coarse mesh, e.g. a \textit{sphere}, and a set of 3D Gaussians, our framework iteratively refines this representation by minimizing the photometric error between the rendered and the ground-truth images. This enables the optimization of both the scene geometry, represented by the underlying mesh, and the appearance defined by the 3D Gaussian representation. We adopt the differentiable rasterization pipeline from \cite{3dgs}, including a custom CUDA kernel for efficient forward and backward passes.

Additionally, we leverage recent advancements in 3DGS-based surface reconstruction, specifically the use of surface elements or \textit{surfels} \cite{dai2024high}. Surfels, similar to 3DGS, are geometric primitives associated with a 2D covariance matrix. In contrast to \cite{dai2024high, sugar} we simultaneously optimize Gaussians constrained to the mesh while clamping the covariance in the normal direction to a tiny constant, rather than learning unconstrained surfels and subsequently inferring a mesh.

The precise objective function we optimize can be varied to suit the specific task, and we typically include a weighted sum of terms from three broad families: \newline
\textit{Photometric losses}: We use $L_1$ or $L_2$ losses between predicted and ground truth RGB pixel values to supervise the optimization of the 3D Gaussians. \newline
\textit{Geometric regularization}: Explicitly optimizing geometry jointly with appearance, we can incorporate Laplacian regularization $L_{LL}$ which penalizes deviations of a vertex from the centroid of its nearest neighbors, thus promoting a smooth surface. We can also use estimates of geometry from other models, e.g. surface normals \cite{garcia2025finetuningimageconditionaldiffusionmodels}.\newline
\textit{Object segmentation}: An $L_2$ silhouette masking loss. This loss compares the predicted silhouette, obtained by adapting the 3D Gaussian rasterizer to object-identity values, against a ground truth object mask obtained using SAM2 \cite{sam2}. However, the binary nature of the ground-truth mask presents a challenge as non-overlapping regions provide no gradient information. To address this, we smooth the binary mask with the Euclidean Distance Transform. This smoothing ensures that gradients can propagate throughout the image, even in areas where the predicted and ground-truth silhouettes do not initially overlap.

\section{Results}
To highlight the contribution of our proposed general framework, we demonstrate its capabilities in novel-view synthesis, geometry reconstruction, and 3D asset generation on two datasets:

\textit{Simulation}: A synthetic dataset, generated using the YCB objects \cite{calli2015benchmarking}, consists of 50 posed images for each of 64 objects.  This dataset was divided 80\%/20\% into train and test sets.\newline %
\textit{Real-to-sim}: A novel dataset captured on the ALOHA2 platform \cite{aldaco2024aloha} consisting of 6 observation trajectories (approx. 800 frames in total) including multiview RGB from 4 cameras together with recorded joint angles. 16 frames from the moving cameras were held out for evaluation.

All our experiments are run on a single GPU NVIDIA H100 (80GB VRAM). Please refer to the Appendix for more details.

\subsection{Simulation}
\subsubsection{Geometry reconstruction}

We assess the quality of SplatMesh-based object reconstruction on the Simulated YCB dataset.  Reconstruction quality was evaluated using the Chamfer Distance (CD) computed on 10 000 points sampled uniformly on both the ground truth and reconstructed meshes.  Our full framework obtains $\mathrm{CD}=\mathbf{0.073} \ \mathrm{mm}^2$.  Without Laplacian mesh regularization we see increased geometric error ($\mathrm{CD}=0.237 \ \mathrm{mm}^2$) due to high frequency artifacts.  An alternative ablation without the surface-aligned Gaussians (surfels) constraint obtains $\mathrm{CD}=0.122 \ \mathrm{mm}^2$.

While our method directly optimize a mesh, NeRFacto produces a radiance field without explicit geometry. The meshes we recovered with NeRFacto's Poisson Surface reconstruction pipeline displayed significant artifacts, e.g. floaters and reconstruction of the background, and we were not able to compare them.

\subsubsection{Novel-view synthesis}
\label{sec:nvs}

\begin{figure}
    \centerline{\includegraphics[width=0.6\columnwidth]{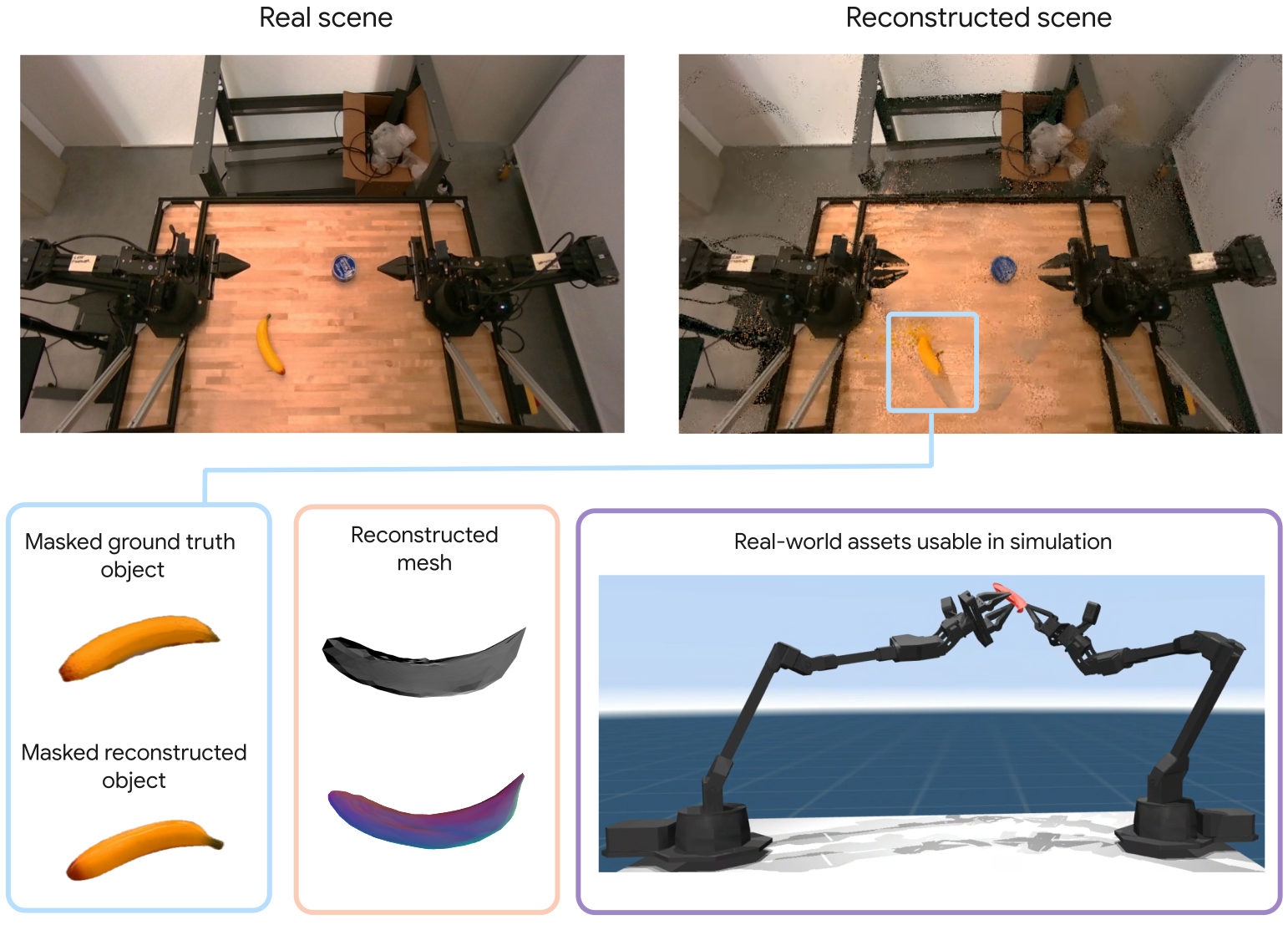}}
    \caption{Recovering real world assets from robot data only.}
    \label{fig:banana-mpc-full}
\end{figure}

We benchmark our method in simulation against two radiance-field techniques: NeRFacto, a state-of-the-art implementation of NeRF in the nerfstudio framework \cite{nerfstudio}, and 3DGS, also implemented in nerfstudio. The strong performance in novel-view synthesis and fast reconstruction make these approaches well-suited for robotics applications. All methods are optimized for 15000 iterations. Additionally, we conduct an ablation study to investigate the influence of mesh reconstruction regularization terms on novel-view synthesis. Specifically, the performance of the proposed reconstruction method is assessed both with and without the inclusion of Laplacian and average edge length loss functions, and without constraining the covariance. Tab.~\ref{table:nvs-sim} shows our results when compared to benchmarks and ablations. We report the standard photometric quality metrics commonly utilized in radiance-field techniques: Peak Signal-to-Noise-Ratio (PSNR), Structural Similarity Index (SSIM), and Learned Perceptual Image Patch Similarity (LPIPS) between the predicted and ground truth images across all YCB objects.

\begin{wraptable}{r}{0.5\textwidth} %

    \centering 
    \caption{Novel-view synthesis metrics on the simulated YCB object dataset after 15K iterations.}
    \resizebox{\linewidth}{!}{
    \begin{tabular}{l c c c}
        \toprule
        
         & PSNR $\uparrow$ & SSIM $\uparrow$ & LPIPS $\downarrow$ \\
        
        \midrule 
        
        NeRFacto & 30.29 & 0.961 & 0.057 \\
        3DGS & 26.97 & \textbf{0.972} & 0.083 \\
        
        \midrule 

        Ours & \textbf{30.91} & 0.970 & \textbf{0.044}\\
        Ours w/o mesh reg. & 30.70 & 0.970 & 0.045  \\
        Ours w/o surfels & 25.82 & 0.954 & 0.071 \\

        \bottomrule 
    \end{tabular}
    }
    \label{table:nvs-sim}
\end{wraptable}

Unlike NeRF, which often exhibits artifacts like "floaters" due to unconstrained volumetric density, our method bounds Gaussians to desired surfaces, leading to higher photometric quality. 3DGS relies on iterative heuristic procedures, such as pruning, cloning, and splitting Gaussians, which necessitate careful tuning and often require several iterations. In contrast, our method benefits from a streamlined optimization process, achieving high-quality reconstructions within a fixed budget of 15000 iterations (~3-4 minutes). The inclusion of mesh regularization terms results in an increased visual quality, with a PSNR of 30.91, compared to a PSNR of 30.70 when regularization was not applied. Without regularization, a uniform weight ($\lambda_{L1}=0.1$) was assigned to the reconstruction loss. When regularization terms were incorporated, object-specific weighting ($\lambda_{LL} \in [0.1, 1.0]$ and $\lambda_{E} \in [0.01, 0.1]$) are employed to accommodate varying levels of guidance required by different objects. Additionally, an ablation study on surfel constraints shows the importance of restricting the 3D Gaussian covariance matrices. Unconstrained Gaussians can expand arbitrarily in 3D space, artificially achieving desired colors or opacities by adopting background colors or near-full transparency. This behavior, while potentially leading to overfitting on training views, results in poor generalization to unseen views.

\subsection{Real}

\begin{table}[htbp] %
\centering %
\caption{Real object recovery metrics for individual YCB prop objects. }
\label{tab:comparison}
\begin{tabular}{lccccc} %
\toprule
& \multicolumn{3}{c}{Geometry $(\sqrt{\mathrm{CD}}, \mathrm{mm}) \downarrow$} & \multicolumn{2}{c}{Novel view synthesis (PSNR, dB) $\uparrow$} \\
\cmidrule(lr){2-4} \cmidrule(lr){5-6} %
Object name & Proprio-only & Aligned TRELLIS & \bf{Ours} & Proprio-only  & \bf{Ours} \\
\midrule
Banana & 11.67 & 10.43 & \bf{7.35} & 16.96 & \bf{24.49} \\
Blue tuna can & 13.92 & 16.17 & \bf{3.31} & 20.17 & \bf{25.45} \\
Lemon & 17.55 & 17.89 & \bf{5.03} & 14.18 & \bf{21.62} \\
Peach & 17.37 & \bf{2.70} & 4.42 & 14.00 & \bf{21.81} \\
Red Apple & 17.16 & 21.40 & \bf{5.09} & 18.22 & \bf{24.94} \\
Strawberry & 18.92 & \bf{1.93} & 3.07 & 16.77 & \bf{22.73} \\
\bottomrule
\end{tabular}
\end{table}

We test the reconstruction of six YCB objects from real robot trajectory data (segmented with text prompts as described in the Appendix). We additionally estimate surface normals for each RGB camera frame independently using the pre-trained model from \cite{garcia2025finetuningimageconditionaldiffusionmodels}.

The objective is a weighted sum of L1 RGB loss, L2 masks smoothed with 2d distance transforms, and L2 between the predicted and estimated surface normals, with mesh Laplacian regularization.   We optimize the mesh vertices, Gaussian parameters, and camera extrinsic rotation parameters.  We run the optimization for 40000 steps, and report in Table \ref{tab:comparison} the Chamfer Distance values for geometry reconstruction and PSNR for novel view synthesis on held-out views of the asset. %

We compare with the \emph{Proprio-only} ablation, the same model with the camera extrinsics frozen at the nominal values.  The geometry fails to converge effectively, demonstrating that without further optimization this low-cost robot platform lacks the high precision required for the reconstruction task.  We also compare shape reconstruction against output from TRELLIS, \cite{xiang2025structured3dlatentsscalable}. This pre-trained 3D foundation model produces 3D appearance and geometry from one or few image views. We provide it with the best hand-chosen masked image from each dataset, and did not find that additional views improved performance. The model does not predict metric scales or object poses, so to enable comparison we optimize the SE3 pose and scale of the TRELLIS mesh using privileged information (Chamfer Distance to the ground truth asset mesh), shown as \emph{Aligned TRELLIS} in the Tab. \ref{tab:comparison}.  We observe that while it can sometimes produce high quality shape predictions, on this real robot dataset it can sometimes introduce anisotropic scale distortions, add spurious geometry like additional ground planes, or fail to capture the 3D structure of simple shapes - see the Appendix.

\subsection{3D asset generation}

\begin{figure}[t!]
    \centerline{\includegraphics[width=0.80\columnwidth]{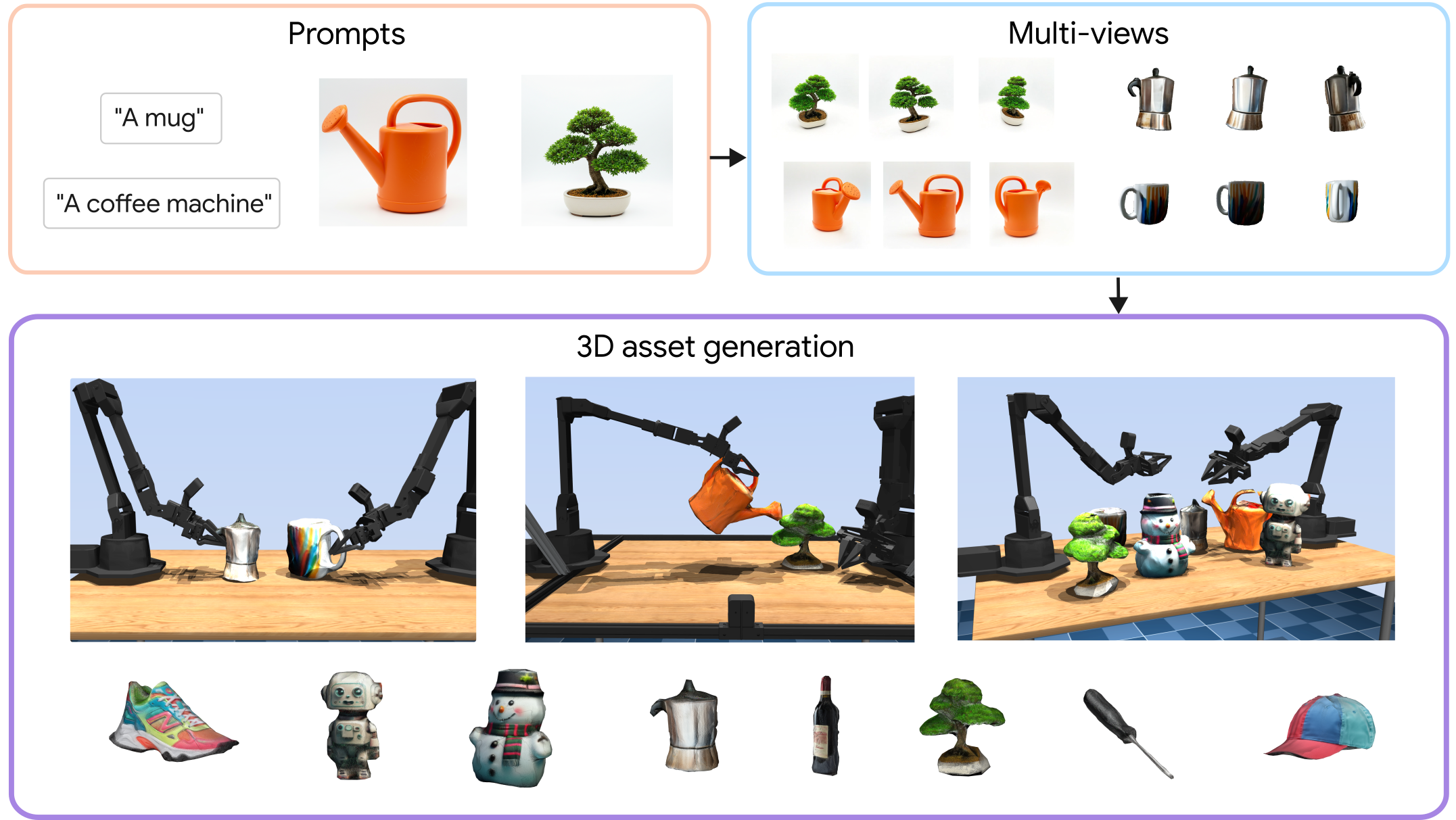}}
    \caption{3D asset generation with SplatMesh. Assets, which are generated from text or a single view image, can be imported in any simulator.}
    \label{fig:genai}
\end{figure}

Our framework extends beyond calibration and object reconstruction to enable the generation of 3D objects from single images or text prompts. The incorporation of single-image or text-based object reconstruction provides a streamlined approach for the rapid creation of diverse object instances with varying geometries and appearances. This capability is particularly valuable for training robust robotic policies by augmenting datasets and enriching synthetic environments.

Given a text prompt or a single image, we leverage CAT3D \cite{gao2024cat3d} to predict geometrically consistent multi-views from single-view input. Then, we follow the procedures outlined in Sec. \ref{sec:splatmesh} for novel-view synthesis and geometry reconstruction using SplatMesh. This approach yields accurate geometry as a mesh and visual appearance as a 3DGS representation. However, as traditional simulators and rendering engines cannot directly render a 3DGS representation, we further optimize a texture map using inverse rendering, supervised on the images produced by our 3DGS differentiable renderer. Fig. \ref{fig:genai} presents a qualitative result of our generated assets within the MuJoCo simulation. Additional results are reported in the Appendix.

\section{Conclusion}

In this paper we explored the the feasibility of tackling a broad range of model identification and scene estimation problems by directly applying end-to-end optimization, fitting standard MuJoCo models to noisy real-world robot data drawn from existing platforms.   We developed a novel explicit 3D scene representation, SplatMesh, which enables gradient signals from RGB pixels to propagate through to arbitrary model parameters including geometry, appearance and camera pose.  We used this framework to refine the calibration of noisy robot and camera poses, and to show the reconstruction of 3D objects from this real data.

This conceptually streamlined but powerful framework is able to leverage existing models to efficiently extract precise information about key simulation quantities from scarce and noisy real world robot data.  It opens a broad spectrum of possibilities for future work enriching and refining physical models with real world data.

\newpage

\section{Limitations}
While our framework demonstrates promising results in reconstructing dynamic robot scenes and novel object geometries, we identify limitations of the present version that motivate future extensions.

Making use of gradient descent to fit parameters is a straightforward way to leverage the differentiability of the model, but is applicable only to smooth parameters and can only provide local information.  In particular this limits meshes learned in this way to those homeomorphic to the topology fixed at initialization (e.g. if the initial geometry is topologically equivalent to a sphere, the refined mesh will also be topologically equivalent to a sphere). This limitation can be mitigated through the choice of initialization mesh structure, but future work will explore more robust and flexible solutions. More generally, gradient descent finds only local minima and so is sensitive to the choice of initialization for non-convex problems.  We can consider several possible ways to handle this:
\begin{itemize}
    \item In our real robot setting, with existing but noisy data, in practice it is often possible to constrain many parameters to a small region of interest for initialization.  
    \item Using more general uncertainty-aware inference methods rather than simply optimization is an interesting direction for future work
    \item Complementing gradient optimization, which excels at high precision and local refinement, with initialization proposed by data-driven learning based approaches
\end{itemize}

Visually, the rendering model used in 3DGS does not enable relighting, so cannot represent effects like reflections and shadows if the dynamic scene elements are moved.  This restriction has been overcome in some later works \cite{saito2024relightable, moenne20243d} but the data gathering requirements may prove a challenge given the real robot constraints.

Finally, since we base our differentiable physics simulation on MuJoCo, we are limited to simulation features supported in its JAX based MJX implementation.  For now this restricts us to rigid objects, although the framework is open source and still under development, so could in principle be extended to support deformable objects as is the case in the C++ Mujoco implementation.

\acknowledgments{Thanks to Nicolas Heess and Norman Di Palo for valuable discussions.}

\bibliography{references}  %

\clearpage
\appendix

\section{Implementation Details}
\label{sec:implementation}

\subsection{Hardware and Software}
All experiments detailed in the main paper were conducted on a single machine equipped with an NVIDIA H100 GPU possessing 80GB of VRAM, using JAX version 0.6.0 and MuJoCo 3.3.2. The baselines were computed using nerfstudio 1.1.5. A single real-world object reconstruction (Section 5.2) over 40,000 steps takes approximately 157 seconds on this machine, after 60s JAX JIT compilation time, and the full experiment takes 3-4 minutes with additional evaluation steps. The optimizaiton loop on the simulated YCB object reconstruction (Section 5.1) over 15,000 steps took less than a minute, and approximately 3-4 minutes for the entire process (dataset generation + optimization) as stated in the main paper.  We used the ALOHA2 model from \cite{menagerie2022github}.

\section{Method Details}
\label{sec:method_details}

\subsection{Scene Representation: SplatMesh}
As described in the main paper, SplatMesh combines an explicit triangle mesh for geometry with 3D Gaussians for appearance.

\textbf{Geometry:} Represented by a triangle mesh (vertices and faces). For these experiments we initialized each object with discretized sphere mesh (642 vertices, 5cm radius). We optimize all 642 vertex translations.  We also added a global 3D translation vector which, although redundant with mesh vertex positions, aids fitting with Adam especially early on when moving the mesh to the object mask.  Mesh connectivity remains fixed throughout optimization.

\textbf{Appearance:} Represented by 3D Gaussians. Each Gaussian is parameterized by:
\begin{itemize}
    \item Barycentric weights ($\mu \in \mathbb{R}^3$): position on the associated triangle face, given as logits of weights for the vertex positions.
    \item Scales ($\mathbb{R}^3$): the covariance $\Sigma \in \mathbb{R}^{3\times 3}$ of each Gaussian is parameterized as face axis-aligned lengths in log scales.
    \item Color: as view-dependent Spherical Harmonics (SH) coefficients, with degree between 0 and 3.
    \item Opacity ($\alpha \in [0, 1]$)
\end{itemize}
\textbf{Coupling:} Gaussians are constrained to the surface of the mesh. We initialize Gaussian means ($\mu$) by sampling barycentric coordinates on the mesh faces, as described in Eq. 1 of the main paper. We found that sampling between 6 and 20 Gaussians per mesh face, while redistributing the particles proportional to face area with stochastic rounding, produces the best result.
The final position and orientation of each Gaussian then evolves through the optimization according to the mesh vertices and barycentric coordinates (see Section 4.2.1).

\textbf{Surface Alignment (Surfel Constraint):} As mentioned in Section 4.3, we adopt a strategy inspired by prior works \cite{dai2024high, Guedon2023-sj}. However, our formulation differs from these methods in that we maintain a mesh at all times, rather than learning unconstrained Gaussians/surfels before inferring a mesh in post-processing. We constrain the Gaussian covariance to be axis aligned with the mesh face normal, and the scale along this normal direction is clamped to a small constant value, \texttt{float32} machine epsilon $\approx 1.2\times 10^{-7}$. This encourages Gaussians to represent surface patches rather than volumetric elements.

\subsection{End-to-End Optimization}
\label{sec:optimization_details}
Our framework utilizes an end-to-end optimization approach based on minimizing prediction error, as outlined in Section 3.1. The core idea is to minimize the discrepancy between real-world observations ($Y$) and simulated observations ($Y' = m(f(s_0, \theta))$) rendered from the current estimate of the scene state and parameters $\theta$. The optimization relies on the differentiability of the entire simulation and rendering pipeline:

\textbf{Differentiable Physics:} We use MuJoCo MJX \cite{brax2021github}, the JAX-native version of the MuJoCo physics simulator \cite{todorov2012mujoco}. This allows us to compute gradients through the physics state ($s = f(s_0, \theta)$), including kinematics (mapping joint angles to Cartesian poses). The physics state is represented by the \texttt{mjx.Data} structure, containing generalized coordinates, velocities, Cartesian poses of bodies/cameras, contact forces, etc. although in the present work we focus on object reconstruction and kinematics.

\textbf{Differentiable Rendering:} We employ the differentiable rasterizer from 3D Gaussian Splatting \cite{3dgs}. This allows gradients from pixel-level losses to flow back to the parameters of the 3D Gaussians (position, covariance, SH coefficients, opacity) and, crucially in our SplatMesh representation, to the underlying mesh vertex positions. We utilize the publicly available implementation based on \cite{3dgs} with a custom interface to JAX.

\subsubsection{Optimization Objectives}
The total loss $L(\theta)$ is a weighted sum of several terms, customized for each task. The parameters $\theta$ include mesh vertex positions, Gaussian parameters, camera poses, robot joint angles, etc.

The losses are defined in terms of rendered image pixel values,  with height, width, and channels $H=480, W=848, C=3$.  We render several different modalities: RGB images $\hat{x}(\theta) \in R^{H\times W\times C}$, using the conventional Gaussian Splatting rasterizer; surface normal images $\hat{n}(\theta)$, where we instead define the RGB colors of each Gaussian to be the X, Y, Z surface normal unit vectors, rescaled to $(0,1)$ and rotated from the world to the camera frame; and mask images $\hat{m}(\theta)$, where we take the colors to be $(1,1,1)$ for all Gaussians and $(0,0,0)$ on the background. We supervise these with corresponding data observations $x$, $n$ and $m$ from datasets described Sec. \ref{sec:datasets} below.

\begin{itemize}
    \item \textbf{Photometric Loss ($\mathcal{L}_\mathrm{photo}$):} The mean absolute difference between rendered RGB images and masked ground truth images:
    \begin{equation}
        \mathcal{L}_{photo}(\theta) = \Vert \hat{x}(\theta) - x \otimes m\Vert_1
    \end{equation}
    \item \textbf{Hard mask Loss ($\mathcal{L}_\mathrm{mask}$):} Encourages the rendered silhouette of an object to match a ground truth segmentation mask. We use an L2 loss on the difference between the rendered segmentation and ground truth binary mask:
        \begin{equation} \mathcal{L}_\mathrm{mask}(\theta) = \Vert \hat{m}(\theta) - m\Vert_2^2
        \end{equation}
    \item \textbf{Soft Mask Loss ($\mathcal{L}_\mathrm{smask}$):}  We apply the squared 2D Euclidean Distance Transform (EDT) as described in Section 4.3 to the ground truth mask $m$ before taking the L2 norm:
        \begin{equation}\mathcal{L}_\mathrm{smask}(\theta) = \Vert \hat{m}(\theta) - [\mathrm{EDT}(m)]^2\Vert_2^2
        \end{equation}
    \item \textbf{Normal Consistency ($\mathcal{L}_{normal}(\theta)$):} Penalizes the difference between predicted surface normals given by the orientations of mesh faces and normals estimated from RGB ground truth.
                \begin{equation}\mathcal{L}_\mathrm{normal}(\theta)= \Vert \hat{n}(\theta) - n\Vert_2^2
                \end{equation}
  
    \item \textbf{Laplacian Smoothing ($\mathcal{L}_\mathrm{laplacian}$):} Encourages smooth surfaces, by penalizing the deviation of each vertex position $v_i(\theta)$ from the average position of its edge-connected neighboring vertices $\mathcal{N}_i$. %
                \begin{equation}\mathcal{L}_{laplacian}(\theta) = \sum_{i} \left\Vert v_i(\theta) - \frac{1}{|\mathcal{N}(i)|} \sum_{j \in \mathcal{N}(i)} v_j(\theta) \right\Vert^2
            \end{equation}
\end{itemize}

We use the Adam optimizer \cite{kingma2014adam}.

\subsubsection{Real shape recovery parameters}

When optimizing real shapes, we did not optimize these but clamped to full opacity, $\alpha=1.0$, finding that this more strongly couples the geometry to the RGB image appearance.  We sampled an average 12 Gaussians per face.  We used Adam learning rates 0.0005 for the Gaussian parameters, 0.0001 for the mesh vertex deltas and camera poses, and 0.001 for the mesh object translation parameters.

See Table \ref{tab:loss_terms_appendix} for a comprehensive breakdown of the loss terms. The target for $\mathcal{L}_\mathrm{normal}$ was predicted surface normals derived from from RGB using \cite{garcia2025finetuningimageconditionaldiffusionmodels}.

\begin{table}[h!] %
    \centering      %
    \caption{Detailed breakdown of the loss terms used in the optimization process. Each term targets a specific aspect of the reconstruction or generation.} %
    \label{tab:loss_terms_appendix} %
    \begin{tabular}{l r} %
        \toprule
        \textbf{Loss term} & \textbf{Weight} \\
        \midrule
        $\mathcal{L}_\mathrm{photo}$          & 1   \\
        $\mathcal{L}_\mathrm{mask}$          & 10            \\
        $\mathcal{L}_\mathrm{smask}$          & $10^{-2}$      \\ %
        $\mathcal{L}_\mathrm{normal}$          & 3             \\
        $\mathcal{L}_\mathrm{laplacian}$ & 3            \\ %
        \bottomrule
    \end{tabular}
\end{table}

\section{Dataset Details}
\label{sec:datasets}

\subsection{Simulation Dataset (YCB Objects)}
As mentioned in the main paper, our simulation dataset uses YCB objects \cite{calli2015benchmarking}.
\begin{itemize}
    \item 64 YCB objects were used.
    \item 50 posed images were generated per object, 80\% for training and 20\% for testing
    \item Camera poses were sampled uniformly on the upper hemisphere around the object center at a fixed distance. The fixed distance is automatically computed to be 2.5 times the spatial extent of the geometry.
    \item Images rendered using MuJoCo's native EGL renderer at 512 $\times$ 512 resolution. Ground truth segmentation masks were also generated.
\end{itemize}

\subsection{Real-World Dataset (ALOHA 2)}
Data was collected using the ALOHA 2 platform \cite{aloha2team2024aloha2enhancedlowcost}.
\begin{itemize}
    \item Platform: Low-cost, bi-manual manipulator with 6 DoF per arm + 1 DoF grippers (14 DoF total).
    \item Sensors: 4 RealSense D405 cameras (2 fixed cameras positioned overhead and at the front of the table, and one camera on each wrist), proprioceptive joint angle sensors (Dynamixel actuators).  Automatic white balance and gain were disabled for consistency between cameras.  Resolution was 848x480 pixels, downsampled to 5 fps. In this work we did not use depth images, only RGB data.
    \item Data: 6 observation trajectories around 40 seconds each with the arms moving around YCB objects, totalling 1168 frames across all cameras before filtering.
    \item Test data: 8 frames from each of the moving wrist cameras per trajectory were held out for novel-view synthesis evaluation (Section 5.2).
\end{itemize}

\begin{figure}[htb!] %
    \centering %

    \begin{subfigure}{0.45\textwidth}
        \includegraphics[width=\linewidth]{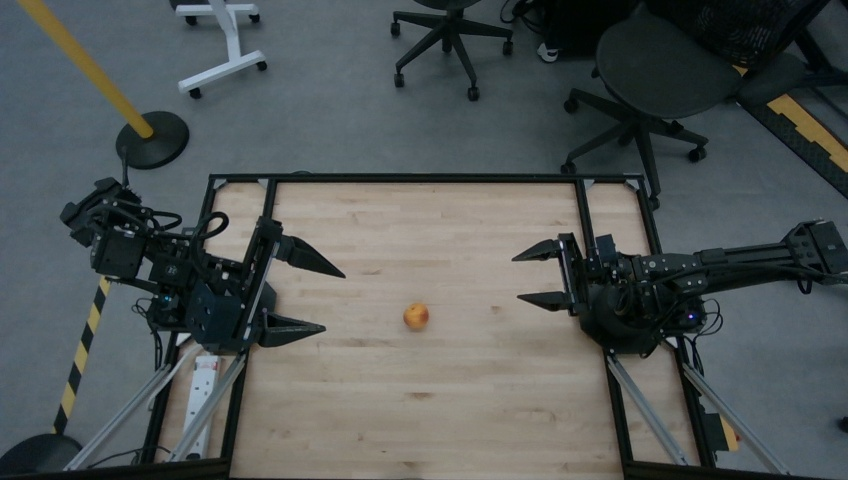}
        \caption{Overhead camera}
    \end{subfigure}\hfill
    \begin{subfigure}{0.45\textwidth}
        \includegraphics[width=\linewidth]{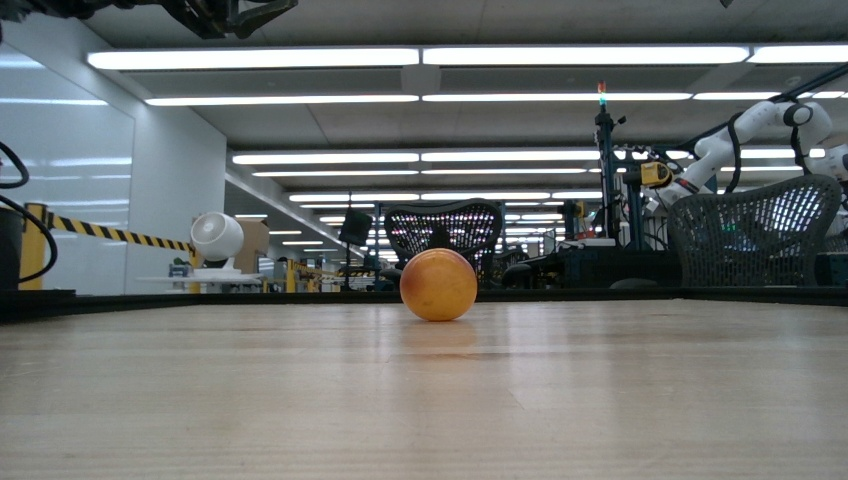}
        \caption{Table camera}
    \end{subfigure}\hfill
    \begin{subfigure}{0.45\textwidth}
        \includegraphics[width=\linewidth]{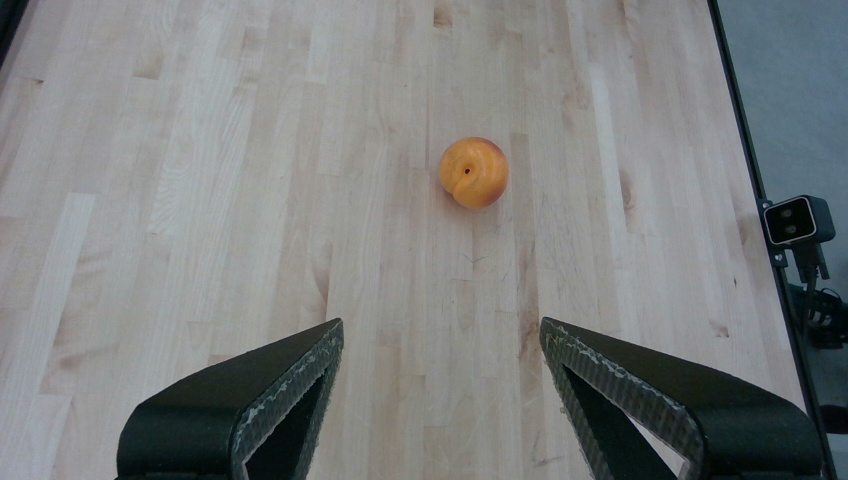}
        \caption{Left wrist camera}
    \end{subfigure}\hfill    
    \begin{subfigure}{0.45\textwidth}
        \includegraphics[width=\linewidth]{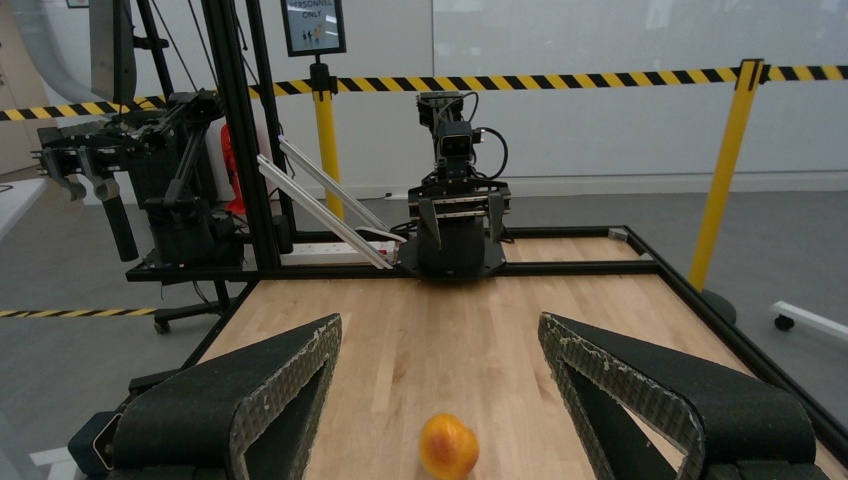}
        \caption{Right wrist camera}
    \end{subfigure}
    \caption{Example images from dataset.}
    \label{fig:dataset-rgb-example}
\end{figure}

\subsubsection{Real-World Object Segmentation Pipeline}
\label{sec:segmentation_details}
To obtain object masks for the real-world YCB objects (Section 5.2), we followed this procedure:
\begin{enumerate}
\item \textbf{Text Prompting:} For each target object (e.g., "banana", "blue tuna can"), a textual description was used as a prompt.
\item \textbf{Bounding Box Proposal:} We used OWL-ViT \cite{minderer2023scaling} to generate bounding box proposals conditioned on the text prompt for each camera frame. The model outputs bounding boxes and associated confidence scores, and we discarded all but the highest confidence box.
\item \textbf{Frame Filtering:} Frames where the maximum detection confidence score for the target object fell below 0.01 were discarded.
\item \textbf{Instance Segmentation:} The selected bounding box was passed as a prompt to SAM2 \cite{sam2} in multimask mode, generating three candidate segmentation masks within that box.
\item \textbf{Mask Postprocessing:} The SAM2 masks were averaged together pixel-wise, and thresholded at 0.5 to obtain binary masks.
\end{enumerate}
This pipeline provided the ground truth masks used for the mask loss ($L_{mask}$) during real-world object reconstruction.

\begin{figure}[t!]
    \centering
    \includegraphics[width=1\columnwidth]{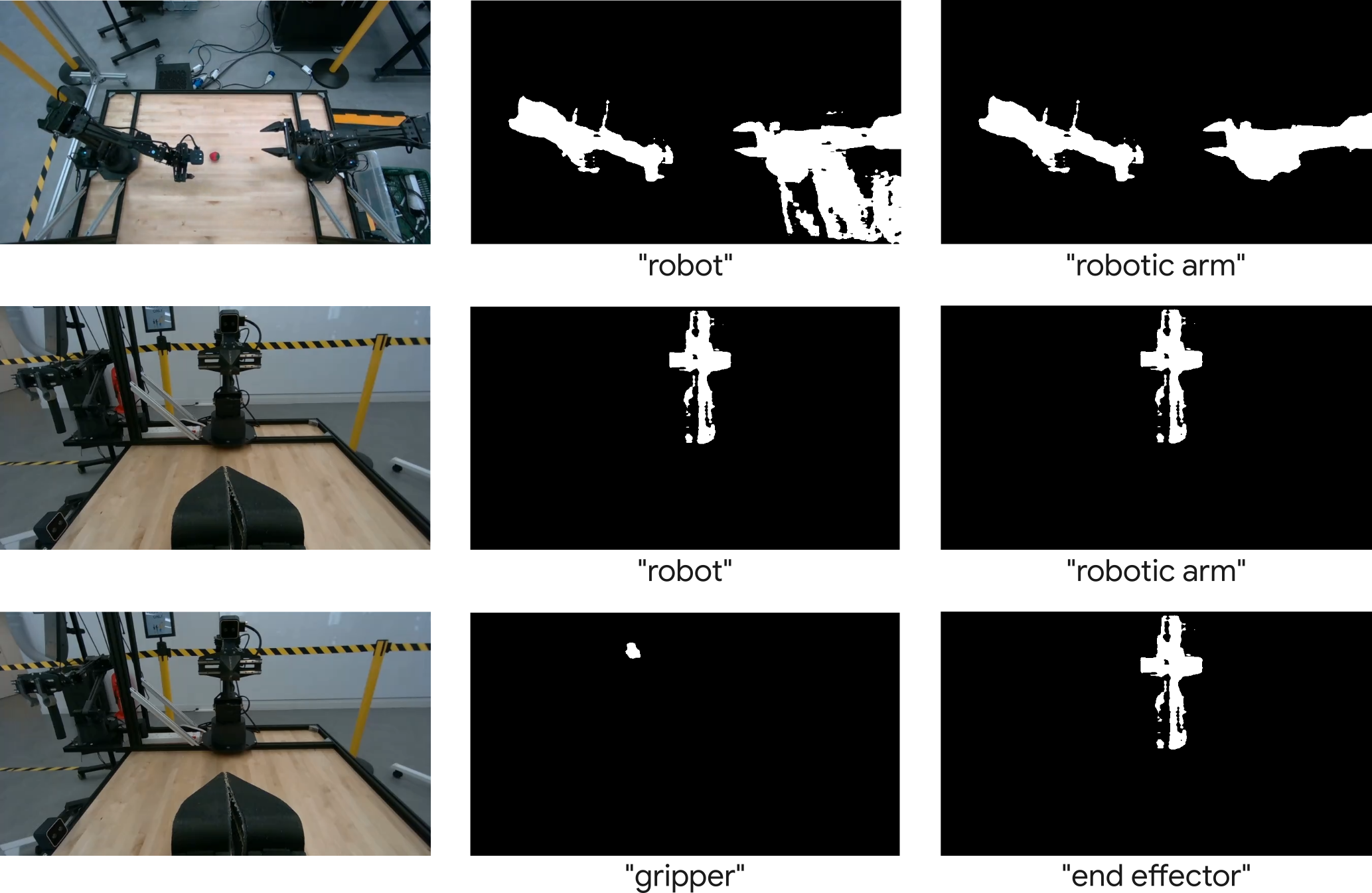}
    \caption{Examples of segmentation difficulties on the ALOHA 2 robot arm using text prompts ("robot arm", "gripper") with OWL-ViT2 for bounding box proposals followed by SAM2 for segmentation. The labels below indicate the OWL-ViT2 proposal text. Masks often miss parts (especially grippers), include background elements, or are inconsistent across views and poses due to lack of distinct texture, uniform color, and complex articulation.}
    \label{fig:masks}
\end{figure}

\subsubsection{Segmentation Challenges on Robot Hardware}
\label{sec:segmentation_challenges}
As stated in the main paper (Section 3.2), standard segmentation models face difficulties with robot hardware like the ALOHA 2 arms, especially compared to distinct semantic objects. Figure \ref{fig:masks} shows examples of semantic segmentation obtained following the approach outlined in \ref{sec:segmentation_details}

The are multiple reasons for the poor results obtained in segmenting our robot platform. First, large parts of the robot arm have simple, uniform coloration, making it hard to distinguish from similarly colored backgrounds or other objects based solely on appearance. Moreover, the robot is an articulated object and its shape changes drastically with its articulation poses, requiring robustness to a wide range of configurations. Grippers, in particular, are often small, complex, and change appearance significantly when opening/closing. This makes mask-based supervision for the robot itself unreliable, motivating our approach of using a known (but potentially inaccurate) robot model and refining its pose via visual feedback from the entire scene, rather than relying on direct segmentation of the robot.

\section{Additional Experimental Results}
\label{sec:additional_results}

\subsection{Robot Calibration}
\label{sec:robot_calibration}
To validate the calibration capability, we performed simulation-to-simulation experiments. We rendered synthetic observations from the ALOHA2 MuJoCo model \cite{menagerie2022github} with the standard OpenGL renderer and took these images and joint angles as ground truth. Noise sampled from a Gaussian distribution with 0 mean and standard deviation $\sigma \in \{0.005, 0.01, 0.02, 0.03\}$ radians was added to the ground truth joint angles to create a dataset with noisy poses. We test calibration by recovering the true poses, optimizing the joint angles and camera perturbations in our framework using SplatMesh for differentiable rendering, and MJX for kinematics. We alternate two Adam optimization steps:

\begin{itemize}
\item Firstly, keeping the pose fixed, we optimize the color of each geometric primitive in the scene by minimizing the L1 photometric loss with respect to the Gaussian color parameters.
\item Secondly, holding the color constant, we optimize the joint angles and fixed camera pose perturbations to maximize the Structural Similarity Index Measure (SSIM) \cite{wang2004image} between the rendered SplatMesh and the ground truth images. We took a weighted average over the multi-view images in each batch, with weights  $w_{fixed}=1.0$ for the two fixed cameras and $w_{wrist}=0.1$ for the two wrist cameras.
\end{itemize}

\textbf{Evaluation:} We measured the error reduction by comparing the initial pose error (due to added noise) and the final pose error after optimization. Errors were measured as average Euclidean distance between the estimated and ground truth Tool Center Points (TCPs) of the two grippers (mm).

\begin{figure}[ht!]
    \centering
    \includegraphics[width=1\columnwidth]{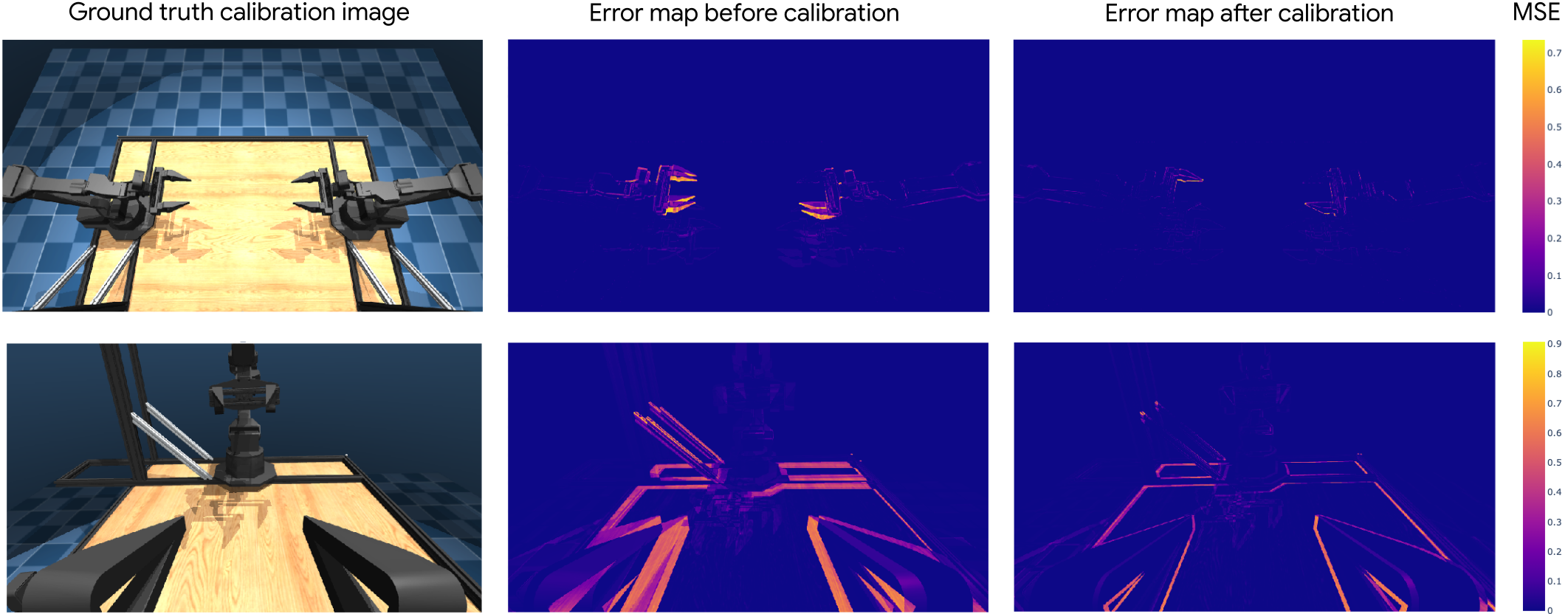}
    \caption{Effect of calibration on wrist camera view for a sample pose with 0.01 radians noise. (Left) Reference image (ground truth rendering). (Center) Pixel-wise MSE error map between the rendering from the noisy pose (using estimated flat colors) and the reference. (Right) Pixel-wise MSE error map between the rendering from the optimized pose and the reference. Brighter colors indicate larger errors. Calibration significantly reduces the visual discrepancy.}
    \label{fig:wrist-cam-calibration}
\end{figure}

\begin{figure}[ht!]
    \centering
    \includegraphics[width=0.5\columnwidth]{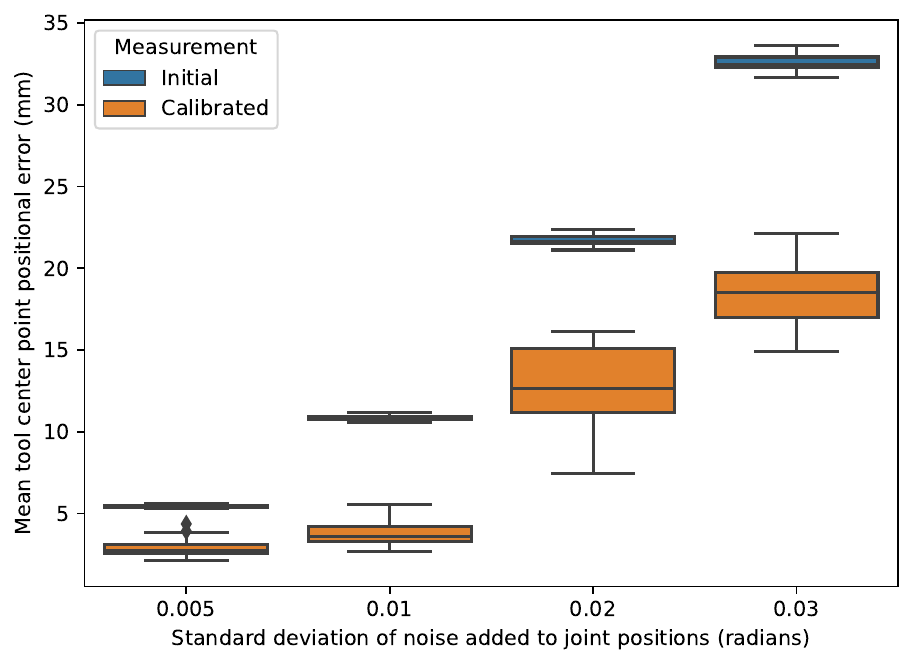}
    \caption{Distribution of TCP errors (mm) before (Initial) and after (Calibrated) optimization for different levels of added joint noise (radians std. dev.). Box plots show median (orange line), quartiles (box), and range (whiskers). Calibration reduces the error across multiple runs with different initializations.}
    \label{fig:calibration-boxplot}
\end{figure}

Results are shown in Tab. \ref{table:calibration} and visually in Figure \ref{fig:wrist-cam-calibration} and Figure \ref{fig:calibration-boxplot}. Our method consistently reduces pose error across different noise levels.

\begin{table}[h!]
    \centering
    \caption{Robot joint pose calibration results. Mean tool pose error (average Euclidean distance between estimated and ground truth TCPs for both arms) before and after optimization via visual feedback.}
    \begin{tabular}{lrr}
    \toprule
    Added noise & Initial mean tool & Final mean tool \\
    (radians std. dev.) & pose error (mm) & pose error (mm) \\
    \midrule
    0.005 & 5.43 & \textbf{2.90} \\
    0.01 & 10.9 & \textbf{3.79} \\
    0.02 & 21.7 & \textbf{12.6} \\
    0.03 & 32.6 & \textbf{18.5} \\
    \bottomrule
    \end{tabular}
    \label{table:calibration}
\end{table}

\subsection{3D Asset Generation Examples}
\label{sec:asset_generation_examples}
\begin{figure}[t!]
    \centering
    \includegraphics[width=\columnwidth]{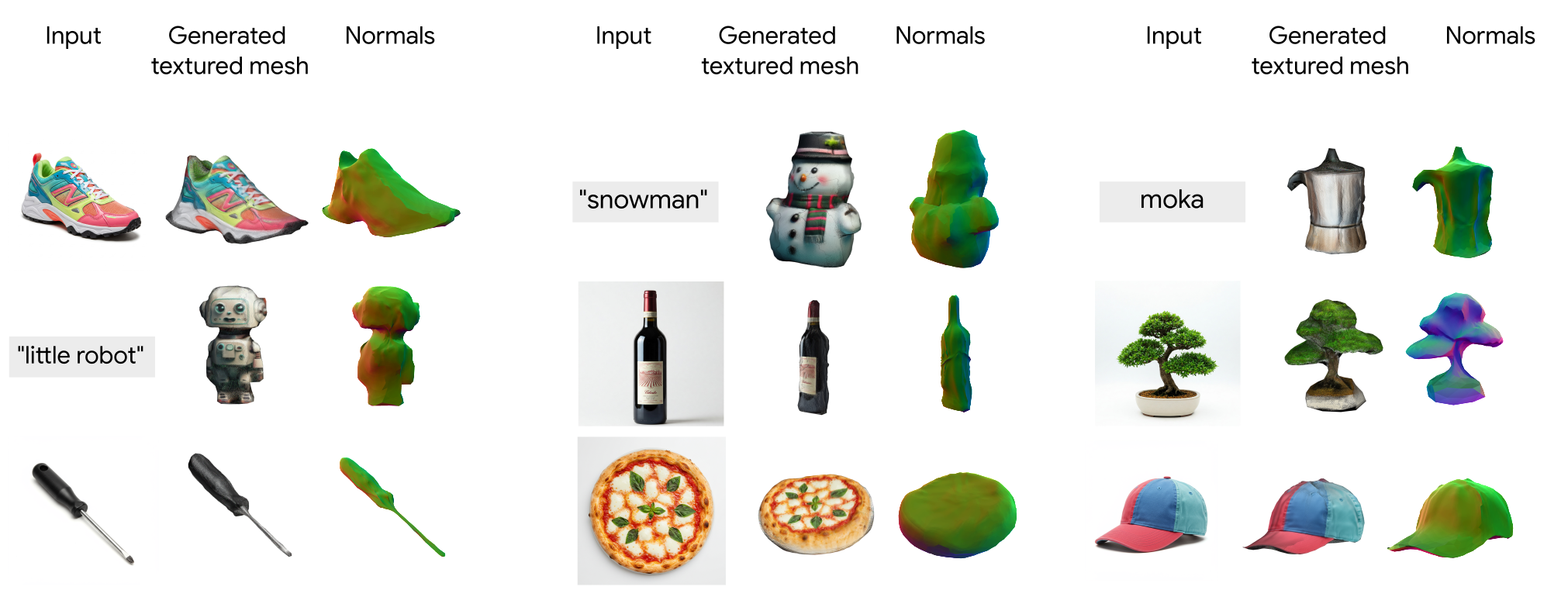} %
    \caption{Additional details on 3D asset generation. For each object: (Left) Input text prompt or single image. (Center) Final textured mesh generated using SplatMesh optimization and texture baking, shown imported into the MuJoCo simulator. (Right) Mesh normals rendered with Trimesh \cite{trimesh}}
    \label{fig:asset_generation_suppl}
\end{figure}

Section 5.3 described our pipeline for generating simulation-ready assets from text prompts or single images using CAT3D \cite{gao2024cat3d} for multi-view generation followed by SplatMesh optimization and texture map baking. Figure \ref{fig:asset_generation_suppl} shows additional examples beyond Figure 4 in the main paper.

The process involves:
\begin{itemize}
    \item Generating consistent multi-view images from the input prompt using CAT3D. This process generates 40 images from a predefined trajectory which revolves around the object.
    \item Optimizing a SplatMesh model (geometry and Gaussians) to fit these multi-view images using the photometric and geometric losses described in Section \ref{sec:optimization_details}.
    \item Once the SplatMesh is optimized, we render its appearance from multiple viewpoints onto the unwrapped UV coordinates of the optimized mesh using inverse rendering to optimize a texture map. This produces a standard textured mesh usable in conventional simulators like MuJoCo.
\end{itemize}

\subsection{YCB Object Reconstruction}
\begin{figure}[ht!]
    \centering
    \includegraphics[width=1\columnwidth]{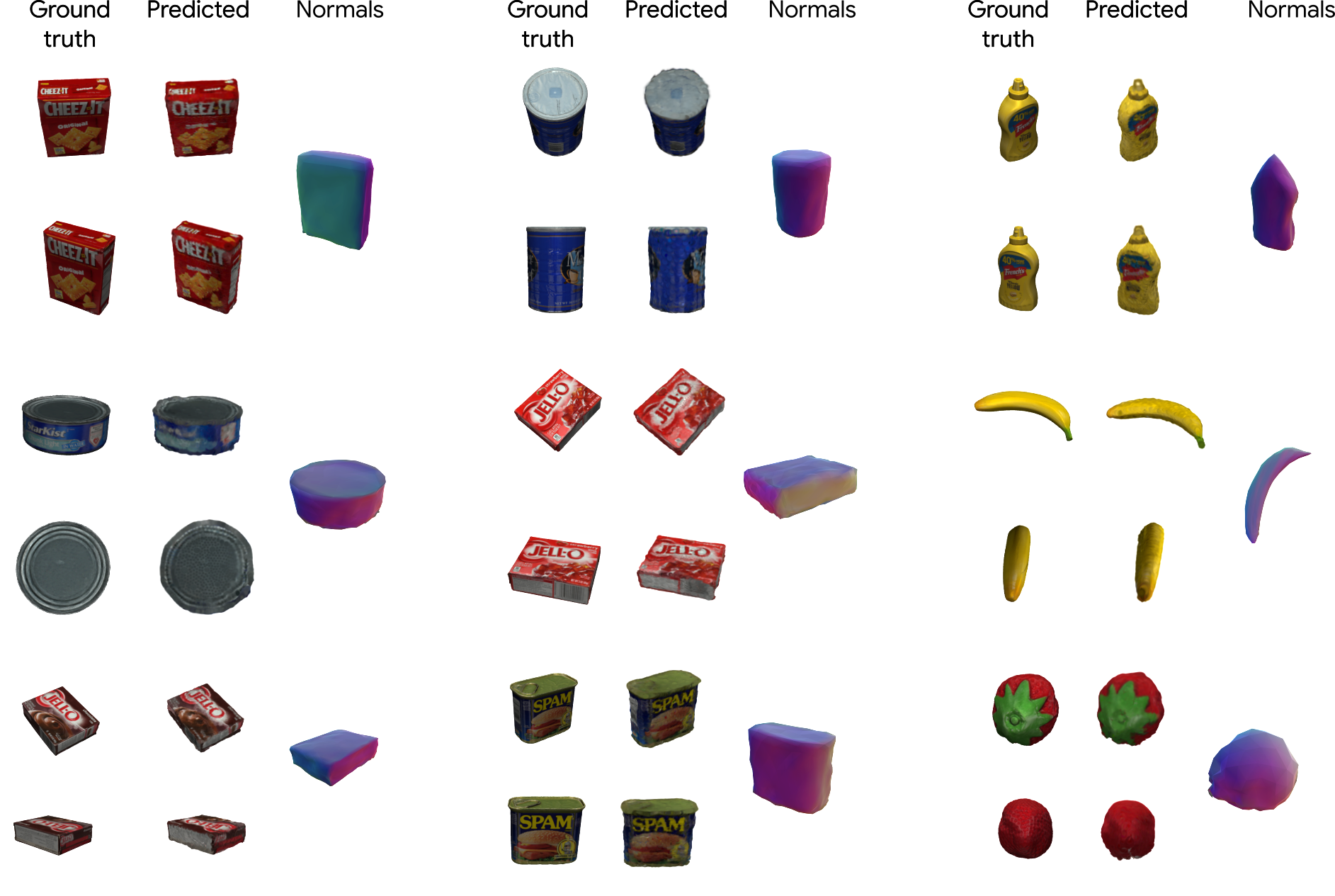}
    \caption{Reconstruction of YCB objects in simulation.}
    \label{fig:ycb}
\end{figure}

Section 5.1.2 in the main paper shows results for novel-viewy synthesis on the YCB dataset. Figure \ref{fig:ycb} shows additional qualitative results on the simulated reconstruction in terms of nove-view synthesis and geometry reconstruction.

\subsection{Real-World Object Reconstruction}
\label{sec:baseline_issues}

We show the six YCB objects reconstructed in the real world and described in the main paper in Figure \ref{fig:all-ycb-grid}.
\begin{figure}[!h] %

    \centering %

    \begin{subfigure}{0.22\textwidth}
        \includegraphics[width=\linewidth]{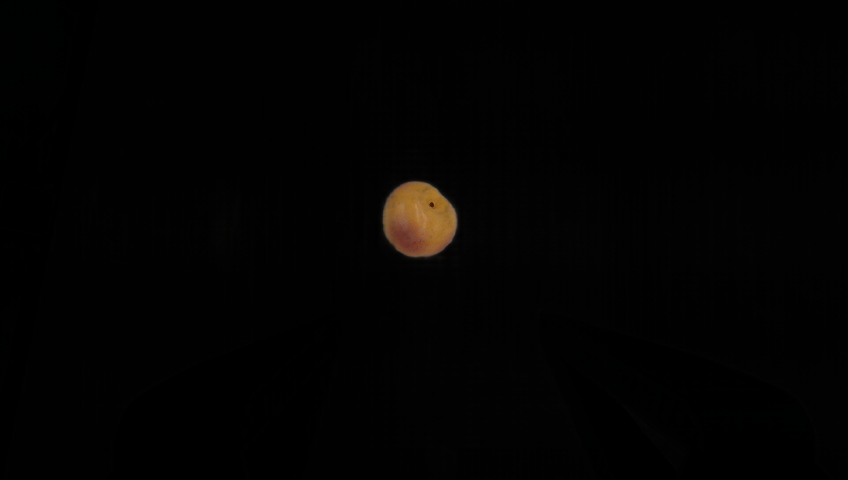}
        \caption{Real RGB}
    \end{subfigure}\hfill
    \begin{subfigure}{0.22\textwidth}
        \includegraphics[width=\linewidth]{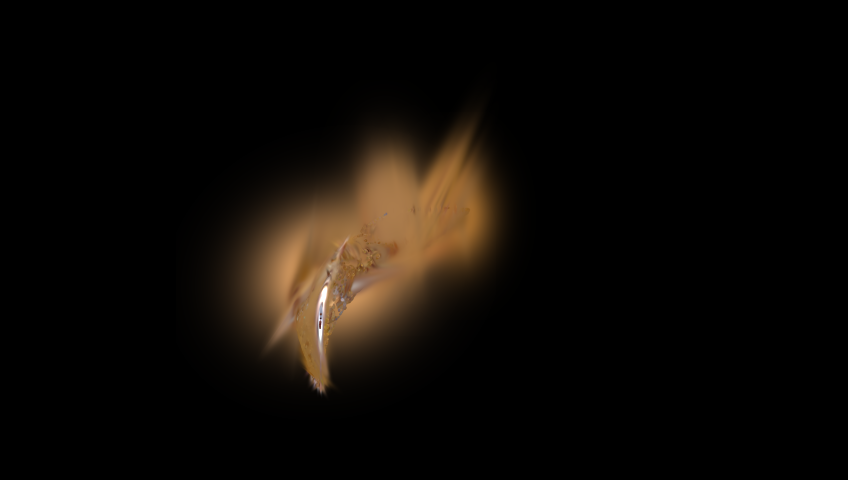}
        \caption{Render RGB}
    \end{subfigure}\hfill
    \begin{subfigure}{0.22\textwidth}
        \includegraphics[width=\linewidth]{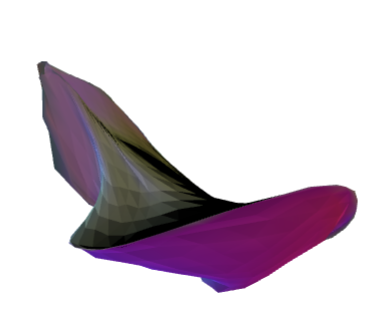}
        \caption{Learned mesh}
    \end{subfigure}    
    \begin{subfigure}{0.22\textwidth}
        \includegraphics[width=\linewidth]{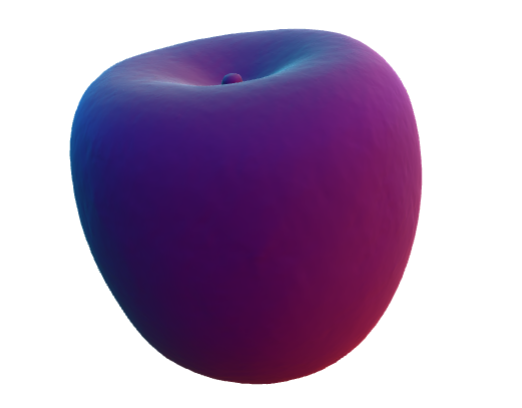}
        \caption{Ground Truth YCB}
    \end{subfigure}
    \caption{Object reconstruction from real robot data fails in the ablation when camera pose is not calibrated jointly with the shape reconstruction.}
    \label{fig:fail}
\end{figure}

\begin{figure}[htb!] %
    \centering %

    \begin{subfigure}{0.22\textwidth}
        \includegraphics[width=\linewidth]{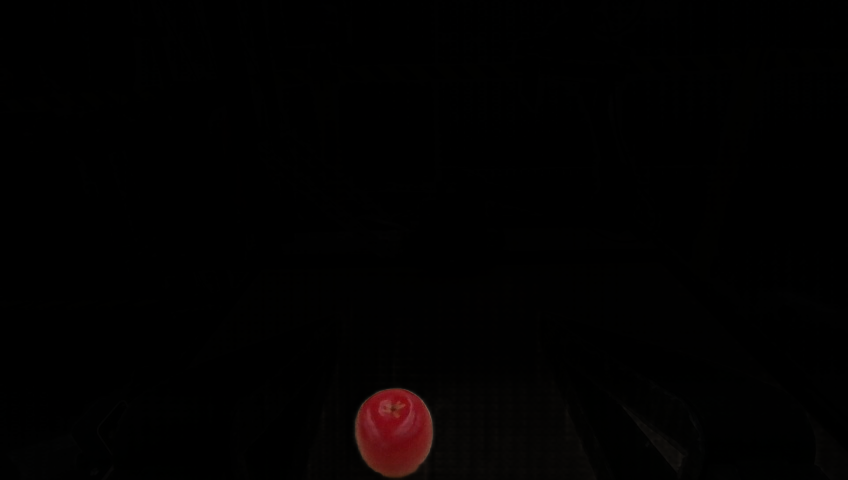}
        \caption{Real RGB}
    \end{subfigure}\hfill
    \begin{subfigure}{0.22\textwidth}
        \includegraphics[width=\linewidth]{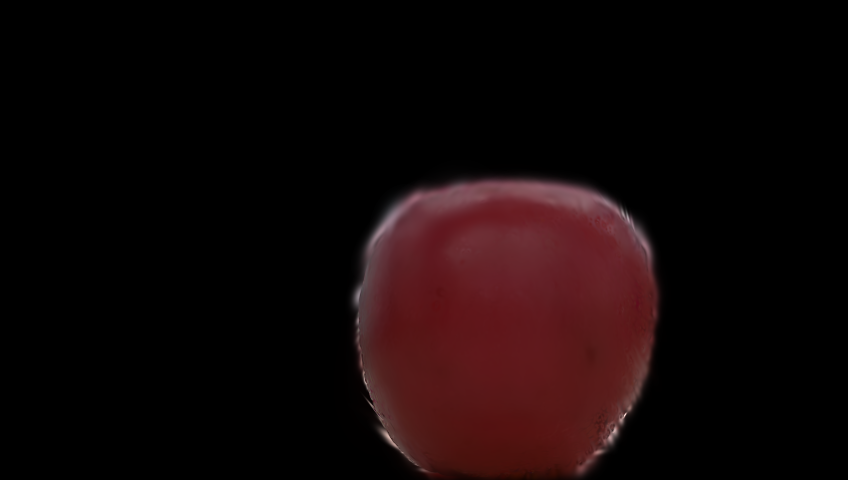}
        \caption{Render RGB}
    \end{subfigure}\hfill
    \begin{subfigure}{0.22\textwidth}
        \includegraphics[width=\linewidth]{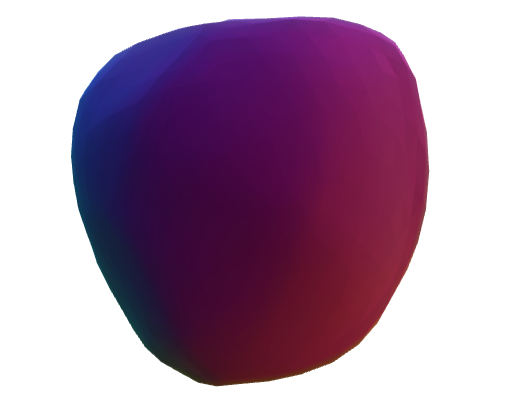}
        \caption{Learned mesh}
    \end{subfigure}    
    \begin{subfigure}{0.22\textwidth}
        \includegraphics[width=\linewidth]{fig/trellis/gt-mesh-apple.png}
        \caption{Ground Truth YCB}
    \end{subfigure}

    \vspace{0.25cm}

    \begin{subfigure}{0.22\textwidth}
        \includegraphics[width=\linewidth]{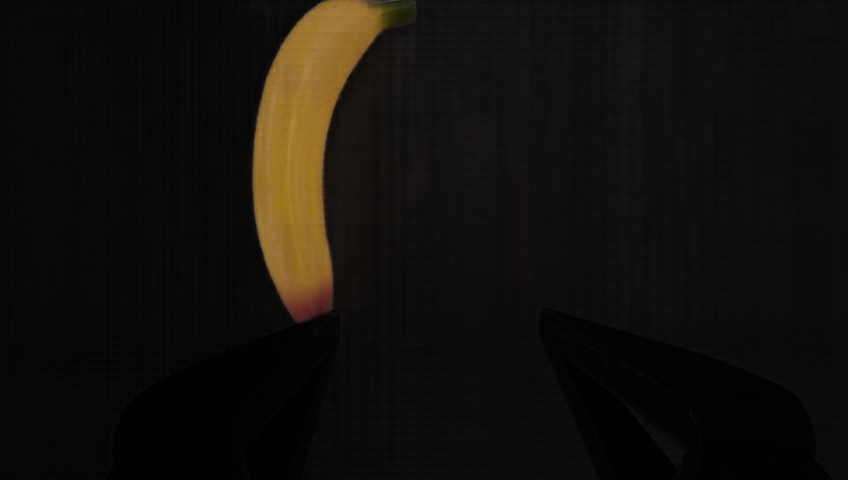}
        \caption{Real RGB}
    \end{subfigure}\hfill
    \begin{subfigure}{0.22\textwidth}
        \includegraphics[width=\linewidth]{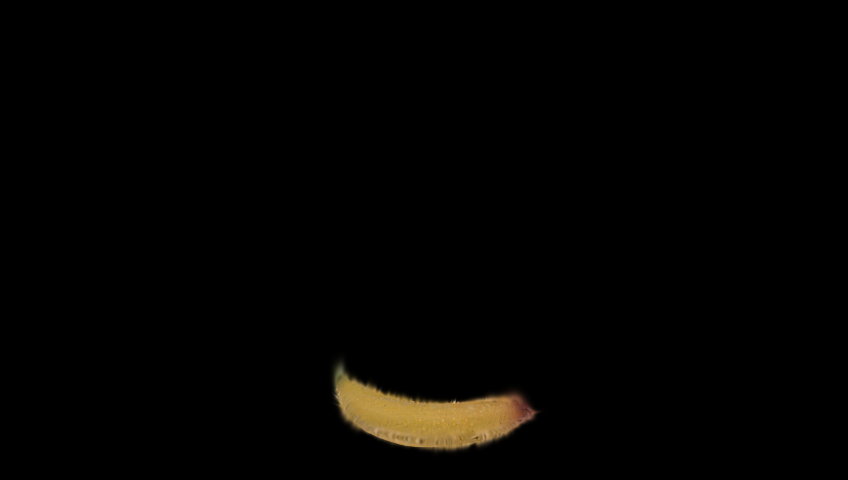}
        \caption{Render RGB}
    \end{subfigure}\hfill
    \begin{subfigure}{0.22\textwidth}
        \includegraphics[width=\linewidth]{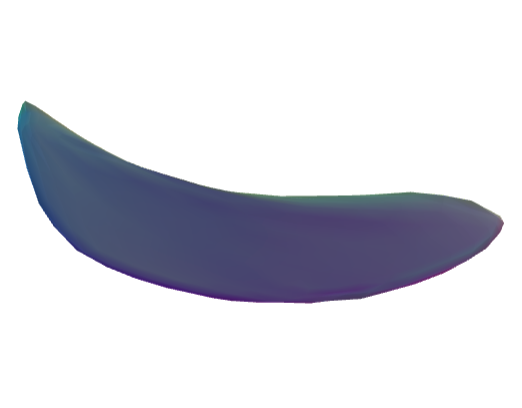}
        \caption{Learned mesh}
    \end{subfigure}    
    \begin{subfigure}{0.22\textwidth}
        \includegraphics[width=\linewidth]{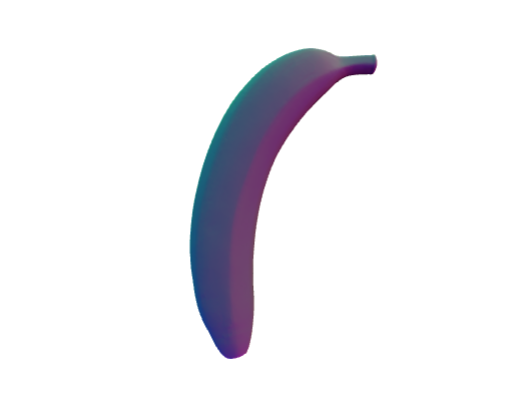}
        \caption{Ground Truth YCB}
    \end{subfigure}

    \vspace{0.25cm}
    \begin{subfigure}{0.22\textwidth}
        \includegraphics[width=\linewidth]{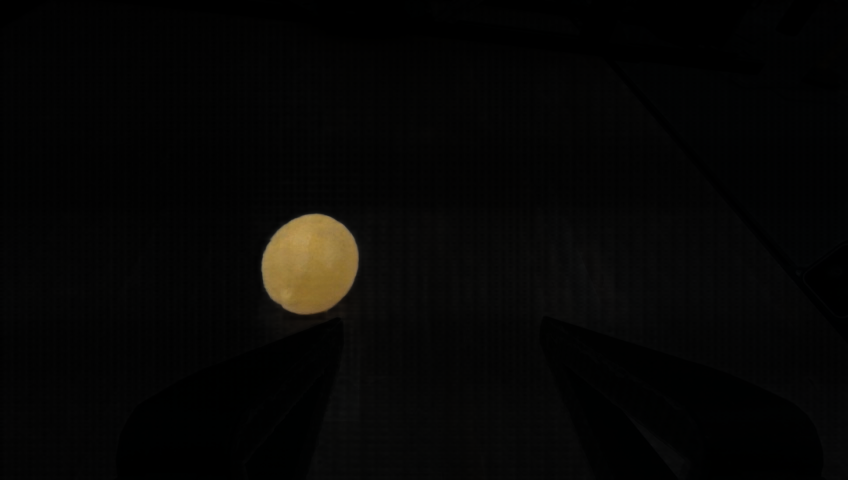}
        \caption{Real RGB}
    \end{subfigure}\hfill
    \begin{subfigure}{0.22\textwidth}
        \includegraphics[width=\linewidth]{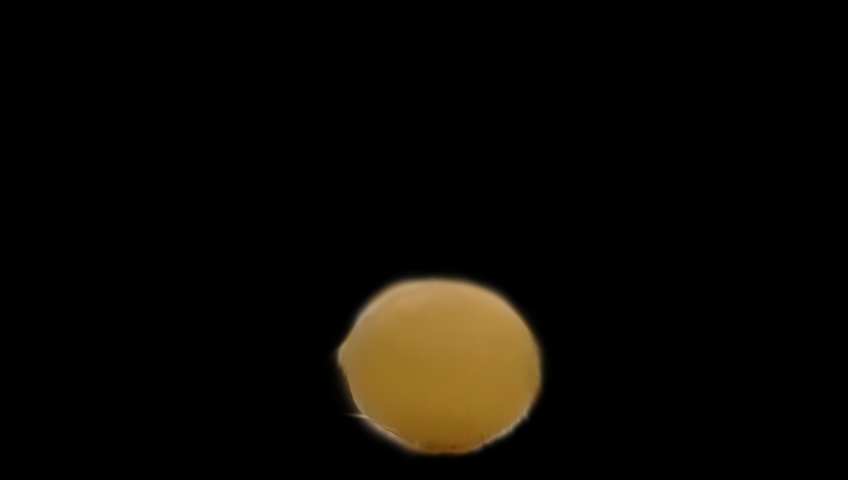}
        \caption{Render RGB}
    \end{subfigure}\hfill
    \begin{subfigure}{0.22\textwidth}
        \includegraphics[width=\linewidth]{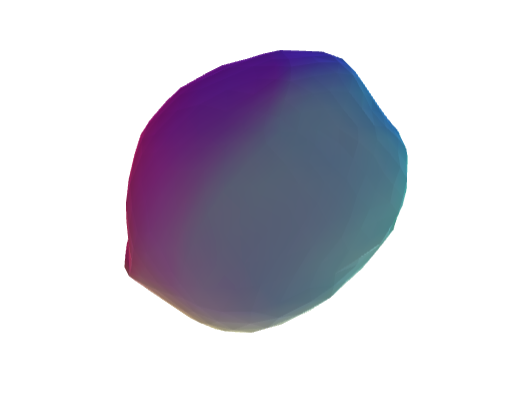}
        \caption{Learned mesh}
    \end{subfigure}    
    \begin{subfigure}{0.22\textwidth}
        \includegraphics[width=\linewidth]{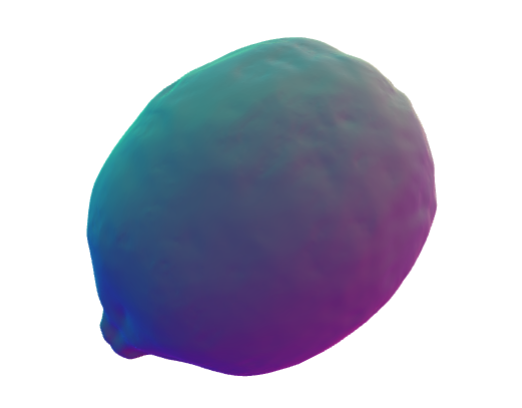}
        \caption{Ground Truth YCB}
    \end{subfigure}

    \vspace{0.25cm}

    \begin{subfigure}{0.22\textwidth}
        \includegraphics[width=\linewidth]{fig/trellis/real-rgb-peach.jpg}
        \caption{Real RGB}
    \end{subfigure}\hfill
    \begin{subfigure}{0.22\textwidth}
        \includegraphics[width=\linewidth]{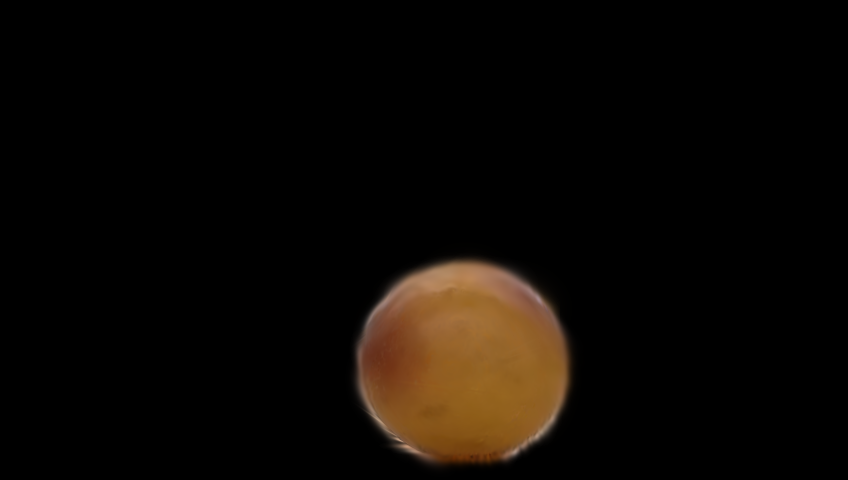}
        \caption{Render RGB}
    \end{subfigure}\hfill
    \begin{subfigure}{0.22\textwidth}
        \includegraphics[width=\linewidth]{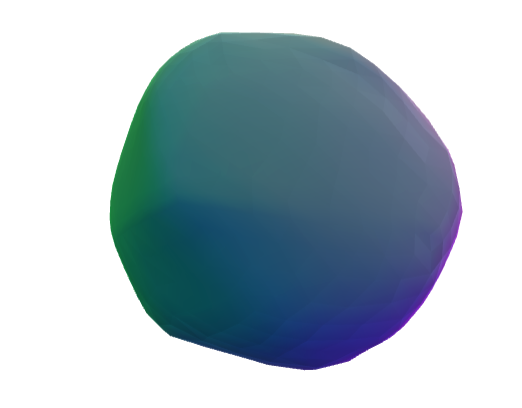}
        \caption{Learned mesh}
    \end{subfigure}    
    \begin{subfigure}{0.22\textwidth}
        \includegraphics[width=\linewidth]{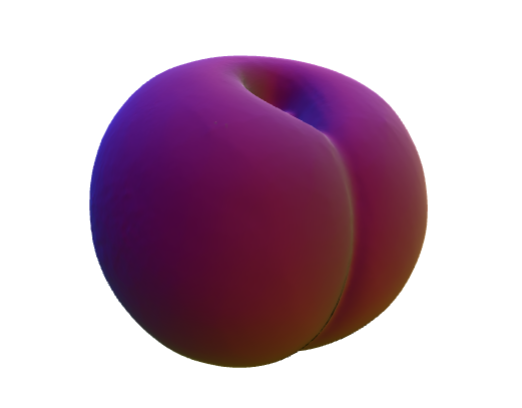}
        \caption{Ground Truth YCB}
    \end{subfigure}

    \vspace{0.25cm}

    \begin{subfigure}{0.22\textwidth}
        \includegraphics[width=\linewidth]{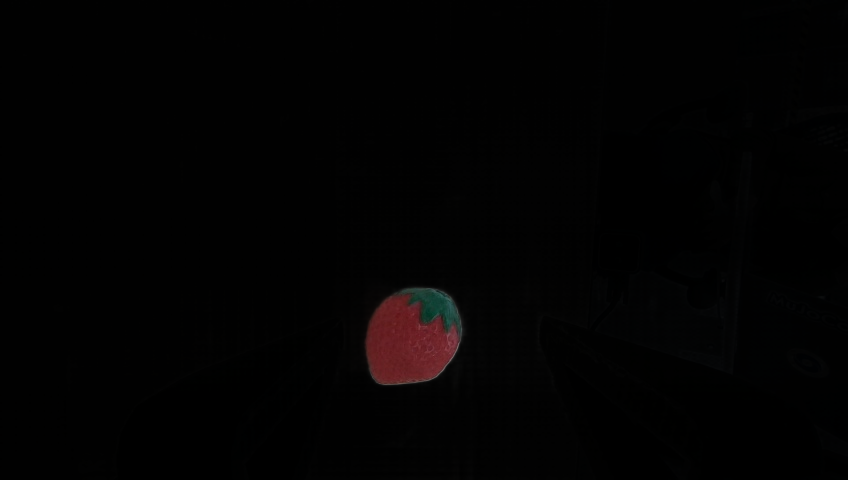}
        \caption{Real RGB}
    \end{subfigure}\hfill
    \begin{subfigure}{0.22\textwidth}
        \includegraphics[width=\linewidth]{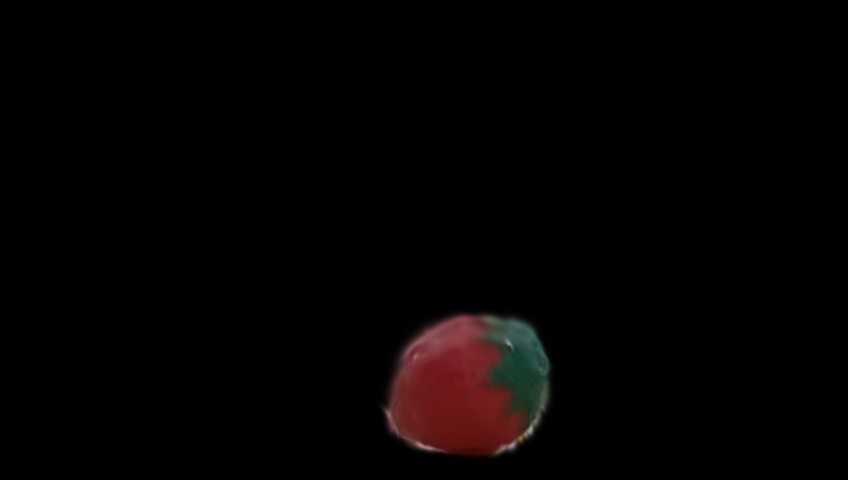}
        \caption{Render RGB}
    \end{subfigure}\hfill
    \begin{subfigure}{0.22\textwidth}
        \includegraphics[width=\linewidth]{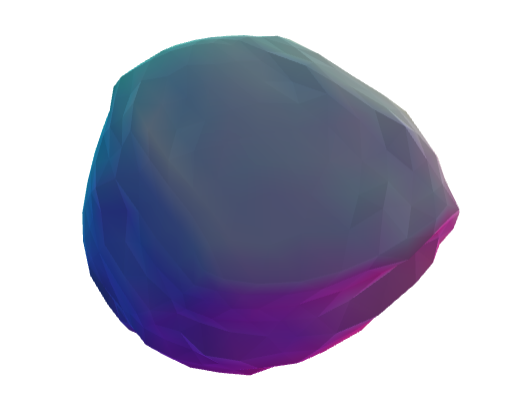}
        \caption{Learned mesh}
    \end{subfigure}    
    \begin{subfigure}{0.22\textwidth}
        \includegraphics[width=\linewidth]{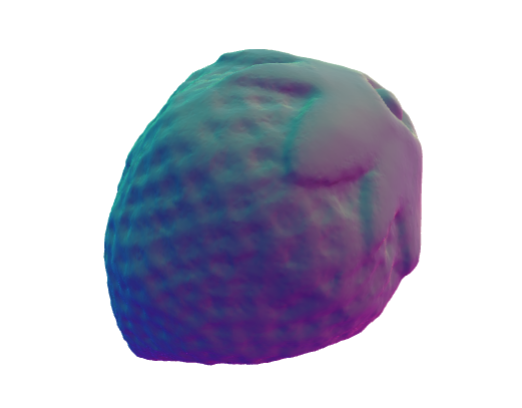}
        \caption{Ground Truth YCB}
    \end{subfigure}

    \vspace{0.25cm}

    \begin{subfigure}{0.22\textwidth}
        \includegraphics[width=\linewidth]{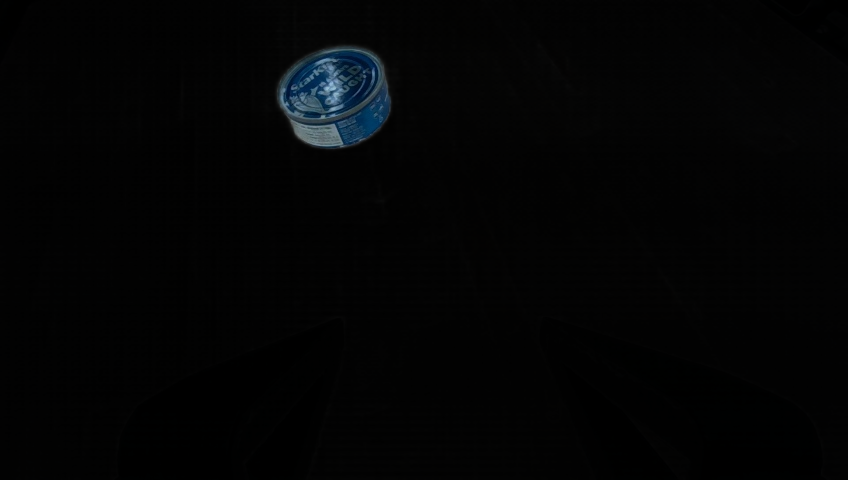}
        \caption{Real RGB}
    \end{subfigure}\hfill
    \begin{subfigure}{0.22\textwidth}
        \includegraphics[width=\linewidth]{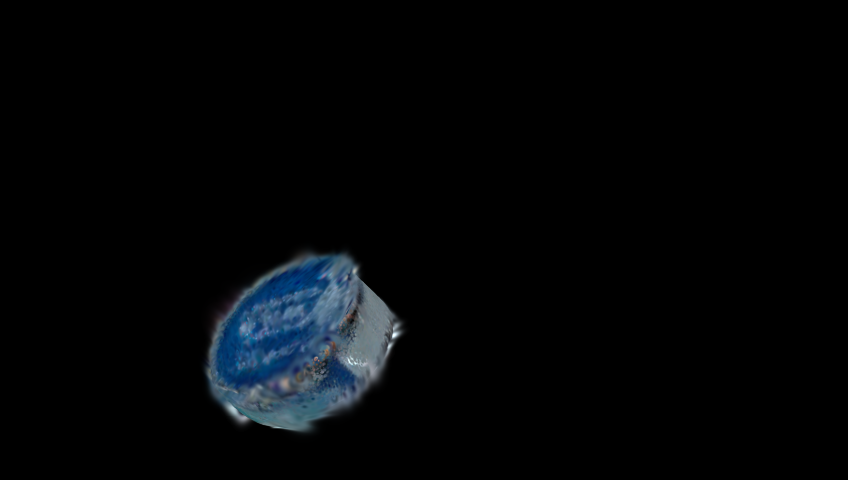}
        \caption{Render RGB}
    \end{subfigure}\hfill
    \begin{subfigure}{0.22\textwidth}
        \includegraphics[width=\linewidth]{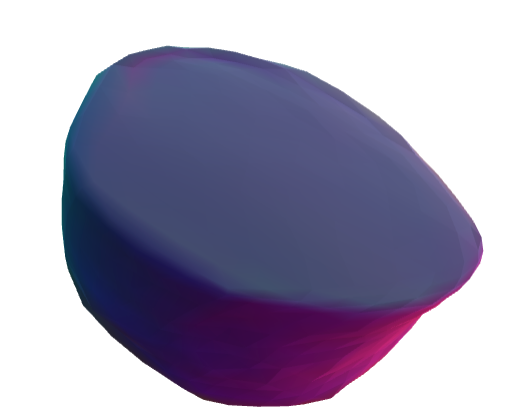}
        \caption{Learned mesh}
    \end{subfigure}    
    \begin{subfigure}{0.22\textwidth}
        \includegraphics[width=\linewidth]{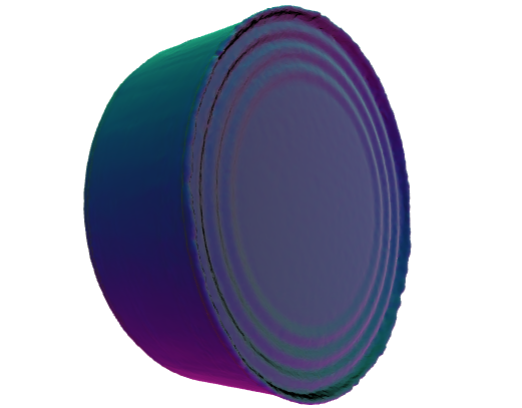}
        \caption{Ground Truth YCB}
    \end{subfigure}

    \vspace{0.25cm}

    \caption{Grid of real and learned YCB object images. Our method recovers metrically accurate posed objects despite the challenging conditions under which the robot operates.}
    \label{fig:all-ycb-grid}
\end{figure}

We also show an example failure mode from the no-calibration ablation in Figure \ref{fig:fail}. Using the ground truth camera poses inferred from the robot's nominal joint angles, it is not possible to correctly reconstruct the image or 3D shape.

Because of these non-negligible errors for the uncalibrated camera poses we additionally note that the PSNR values reported in the novel-view synthesis results in the main text are calculated after an additional optimization step, to align the camera poses for the held-out views. This step happens after training and does not affect the learned textures or geometry: it is done to enable evaluation of the reconstruction for unseen views at our best estimate of the ground truth camera poses.

\begin{figure}[htb!] %
    \centering %

    \begin{subfigure}{0.22\textwidth}
        \includegraphics[width=\linewidth]{fig/trellis/real-rgb-apple.jpg}
        \caption{Real RGB}
    \end{subfigure}\hfill
    \begin{subfigure}{0.22\textwidth}
        \includegraphics[width=\linewidth]{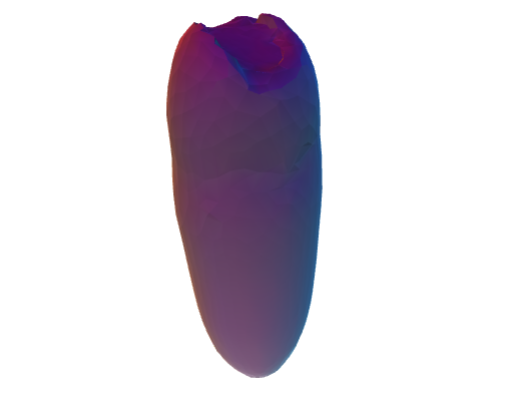}
        \caption{TRELLIS (no scale)}
    \end{subfigure}\hfill
    \begin{subfigure}{0.22\textwidth}
        \includegraphics[width=\linewidth]{fig/trellis/gt-mesh-apple.png}
        \caption{Ground Truth YCB}
    \end{subfigure}
    
    \vspace{0.25cm}
    
    \begin{subfigure}{0.22\textwidth}
        \includegraphics[width=\linewidth]{fig/trellis/real-rgb-banana.jpg}
        \caption{Real RGB}
    \end{subfigure}\hfill
    \begin{subfigure}{0.22\textwidth}
        \includegraphics[width=\linewidth]{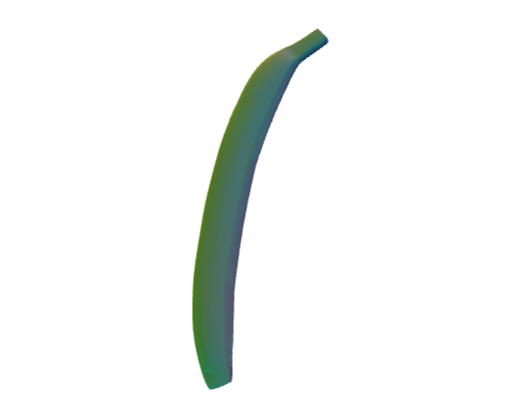}
        \caption{TRELLIS (no scale)}
    \end{subfigure}\hfill
    \begin{subfigure}{0.22\textwidth}
        \includegraphics[width=\linewidth]{fig/trellis/gt-mesh-banana.png}
        \caption{Ground Truth YCB}
    \end{subfigure}

    \vspace{0.25cm}

    \begin{subfigure}{0.22\textwidth}
        \includegraphics[width=\linewidth]{fig/trellis/real-rgb-lemon.jpg}
        \caption{Real RGB}
    \end{subfigure}\hfill    
    \begin{subfigure}{0.22\textwidth}
        \includegraphics[width=\linewidth]{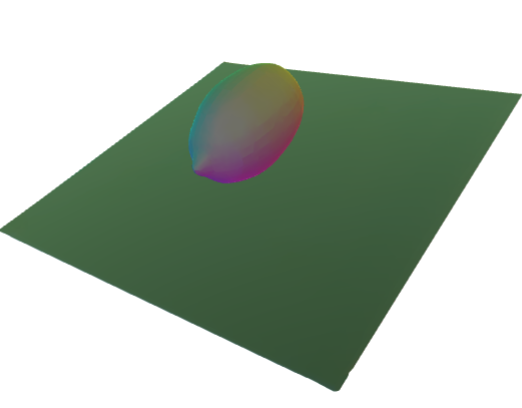}
        \caption{TRELLIS (no scale)}
    \end{subfigure}\hfill        
    \begin{subfigure}{0.22\textwidth}
        \includegraphics[width=\linewidth]{fig/trellis/gt-mesh-lemon.png}
        \caption{Ground Truth YCB}
    \end{subfigure}
    \vspace{0.25cm}

    \begin{subfigure}{0.22\textwidth}
        \includegraphics[width=\linewidth]{fig/trellis/real-rgb-peach.jpg}
        \caption{Real RGB}
    \end{subfigure}\hfill
    \begin{subfigure}{0.22\textwidth}
        \includegraphics[width=\linewidth]{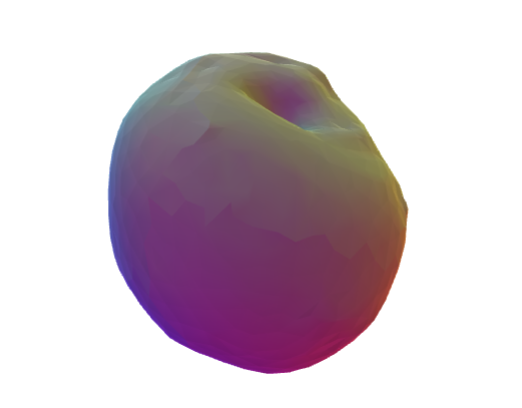}
        \caption{TRELLIS (no scale)}
    \end{subfigure}\hfill
    \begin{subfigure}{0.22\textwidth}
        \includegraphics[width=\linewidth]{fig/trellis/gt-mesh-peach.png}
        \caption{Ground Truth YCB}
    \end{subfigure}
    \vspace{0.25cm} 

    \begin{subfigure}{0.22\textwidth}
        \includegraphics[width=\linewidth]{fig/trellis/real-rgb-strawberry.jpg}
        \caption{Real RGB}
    \end{subfigure}\hfill
        \begin{subfigure}{0.22\textwidth}
        \includegraphics[width=\linewidth]{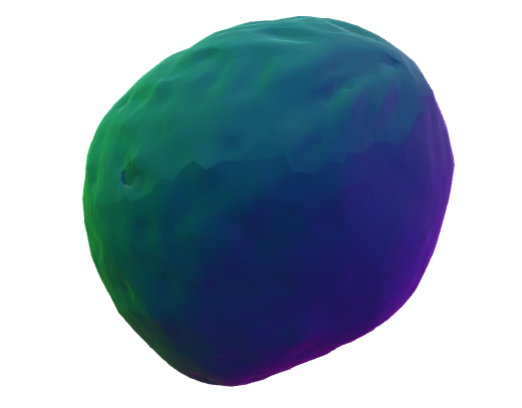}
        \caption{TRELLIS (no scale)}
    \end{subfigure}\hfill
    \begin{subfigure}{0.22\textwidth}
        \includegraphics[width=\linewidth]{fig/trellis/gt-mesh-strawberry.png}
        \caption{Ground Truth YCB}
    \end{subfigure}    

    \vspace{0.25cm}    

    \begin{subfigure}{0.22\textwidth}
        \includegraphics[width=\linewidth]{fig/trellis/real-rgb-tuna.jpg}
        \caption{Real RGB}
    \end{subfigure}\hfill
    \begin{subfigure}{0.22\textwidth}
        \includegraphics[width=\linewidth]{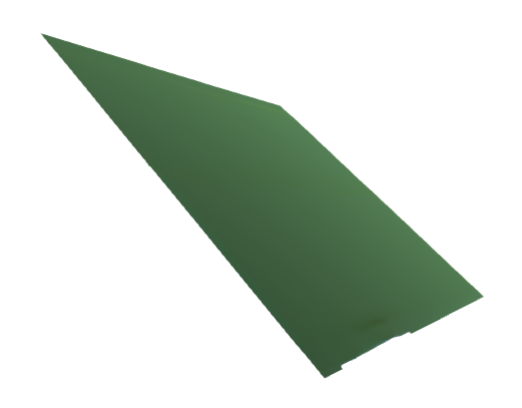}
        \caption{TRELLIS (no scale)}
    \end{subfigure}\hfill
    \begin{subfigure}{0.22\textwidth}
        \includegraphics[width=\linewidth]{fig/trellis/gt-mesh-tuna.png}
        \caption{Ground Truth YCB}        
    \end{subfigure}
    \vspace{0.25cm}

    \caption{Real object image prompts, ground truth meshes and TRELLIS meshes.  While TRELLIS sometimes infers plausible shapes from a single image, it sometimes distorts or includes additional geometry.}
    \label{fig:all-trellis-grid}    
\end{figure}

As mentioned in Section 5.3, we tested the recent 3D reconstruction method TRELLIS \cite{xiang2025structured3dlatentsscalable} on our dataset. This data-driven method is effective at single-view reconstruction and can infer complex unseen geometry based on its training set.  However it exhibited some failure modes on our specific robot data, which may be out of its training distribution (see Figure \ref{fig:all-trellis-grid}). In some cases, the reconstructed object mesh was disproportionately scaled along one axis, not matching the true object shape. Moreover, extra geometry was sometimes generated, such as the additional ground plane in the Lemon example. Finally, For some simple objects such as the Tuna Can, the model failed to capture the basic 3D structure correctly from the provided view(s).

We ran TRELLIS with the default settings.  We manually selected a masked image of each object from our dataset, taking the least occluded, clearest view we could find.  We also tried the experimental multi-view feature in TRELLIS but did not find that this improved the results. Note also that the output from TRELLIS does not give scale or pose for the objects, whereas our method is metrically accurate and positions the object with its 6D pose accurately within the robot workspace.

These issues highlight the difficulty of obtaining accurate, metric-scale geometry directly from limited, potentially noisy views captured by a low-cost robot, motivating our approach that leverages multi-view consistency and end-to-end optimization.

\clearpage

\end{document}